\documentclass[lettersize,journal]{IEEEtran}
\usepackage{amsmath,amsfonts}
\usepackage{algorithmic}
\usepackage{algorithm}
\usepackage{array}
\usepackage[caption=false,font=normalsize,labelfont=sf,textfont=sf]{subfig}
\usepackage{textcomp}
\usepackage{stfloats}
\usepackage{url}
\usepackage{verbatim}
\usepackage{graphicx}
\usepackage{cite}
\usepackage{booktabs}
\usepackage{xcolor}
\usepackage{orcidlink}

\begin{document}

\title{VideoDreamer: Customized Multi-Subject Text-to-Video Generation with Disen-Mix Finetuning on Language-Video Foundation Models}

\author{Hong~Chen\textsuperscript{\orcidlink{0000-0002-0943-2286}},~Xin~Wang\textsuperscript{\orcidlink{0000-0002-0351-2939}},~\IEEEmembership{Member,~IEEE,}~Guanning~Zeng\textsuperscript{\orcidlink{0009-0009-3783-9276}},~Yipeng~Zhang\textsuperscript{\orcidlink{0009-0002-0886-8296}},
~Yuwei~Zhou\textsuperscript{\orcidlink{0000-0001-9582-7331}},~Feilin~Han\textsuperscript{\orcidlink{0000-0001-7463-2252}},~Yaofei~Wu,~and~Wenwu~Zhu\textsuperscript{\orcidlink{0000-0003-2236-9290}},~\IEEEmembership{Fellow,~IEEE}

\thanks{
This work was supported by National Natural Science Foundation of China No. 62222209, Beijing National Research Center for Information Science and Technology under Grant BNR2023TD03006, 
and Beijing Key Lab of Networked Multimedia. \textit{(Corresponding authors: Xin Wang and Wenwu Zhu.)}
}

\thanks{Hong Chen, Xin Wang, Guanning Zeng, Yipeng Zhang, Yuwei Zhou, and Wenwu Zhu are with the Department of Computer Science and Technology, Tsinghua University, Beijing 100084, China (e-mail:\{h-chen20, zgn21, zhang-yp22, zhou-yw21\}@mails.tsinghua.edu.cn, \{xin\_wang, wwzhu\}@tsinghua.edu.cn). Xin Wang and Wenwu Zhu are also with Beijing National Research Center for Information Science and Technology. Feilin Han is with Department of Film and TV Technology at Beijing Film Academy, China. (e-mail:hanfeilin@bfa.edu.cn). Yaofei Wu is with Beijing University of Technology.(23027313@emails.bjut.edu.cn)}
}

\markboth{Preprint.}%
{Shell \MakeLowercase{\textit{et al.}}: A Sample Article Using IEEEtran.cls for IEEE Journals}


\maketitle

\begin{abstract}
Customized text-to-video generation aims to generate text-guided videos with user-given subjects, which has gained increasing attention. However, existing works are primarily limited to single-subject oriented text-to-video generation, leaving the more challenging problem of customized multi-subject generation unexplored. In this paper, we fill this gap and propose a novel VideoDreamer framework, which can generate temporally consistent text-guided videos that faithfully preserve the visual features of the given multiple subjects. Specifically, VideoDreamer adopts the pretrained Stable Diffusion with temporal modules as its base video generator, taking the power of the text-to-image model to generate diversified content. The video generator is further customized for multi-subjects, which leverages the proposed Disen-Mix Finetuning and Human-in-the-Loop Re-finetuning strategy, to tackle the attribute binding problem of multi-subject generation. Additionally, we present a disentangled motion customization strategy to finetune the temporal modules so that we can generate videos with both customized subjects and motions. To evaluate the performance of customized multi-subject text-to-video generation, we introduce the MultiStudioBench benchmark. Extensive experiments demonstrate the remarkable ability of VideoDreamer to generate videos with new content such as new events and backgrounds, tailored to the customized multiple subjects. 
\end{abstract}


\section{Introduction}
\label{sec:intro}
Pretrained on large-scale multimodal datasets~\cite{videodataset1,videodataset2,cogvideo,laion}, text-to-video models~\cite{cogvideo,makeavideo,makeyourvideo,text2video0,free-bloom,imagenvideo,magicvideo} can generate temporal-coherent and photo-realistic videos following the given textual prompts. However, relying solely on textual prompts poses a challenge in precisely controlling the visual details of the generated videos. For instance, when a user desires to create a video of \textit{``their favorite pet dog surfing on the ocean''}, it becomes difficult to determine a textual prompt that indicates the inclusion of a visually similar dog to their own pet. Consequently, customized text-to-video generation~\cite{makeyourvideo,text2video0}, where a video that reflects user-specific concepts is expected to be generated with textual prompts, has received increased attention recently. However, existing customized text-to-video generation works~\cite{guo2023animatediff,text2video0} primarily focus on a single subject, limiting their application to broader scenarios, where a user may want to generate a video of their pet dog and cat playing together, which involves multiple subjects.

\begin{figure*}[htbp]
    \centering
    \includegraphics[width=\linewidth]{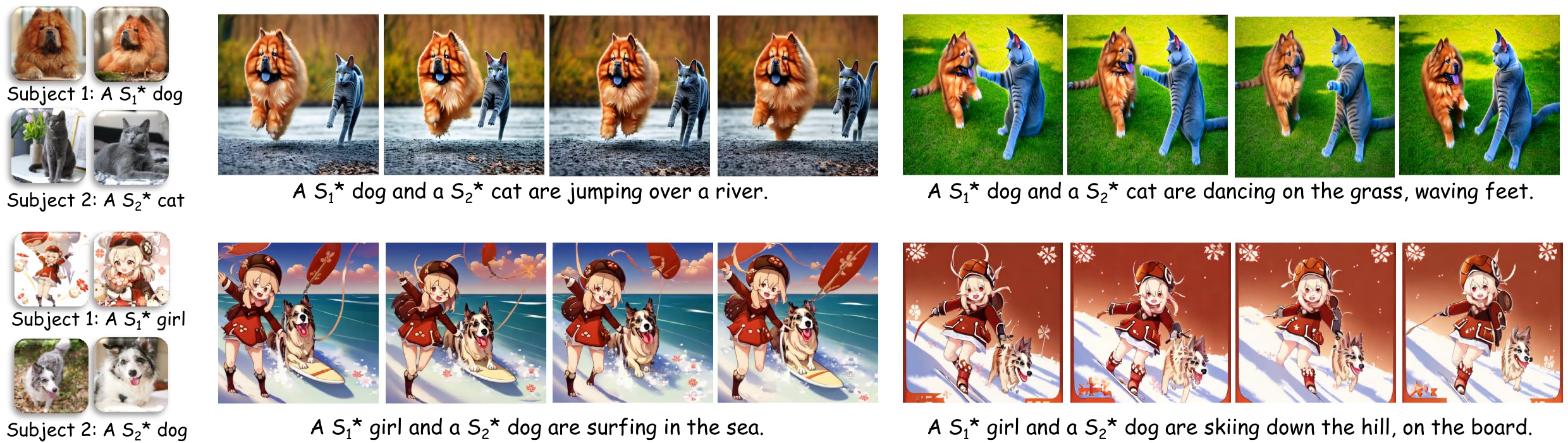}
    \caption{Customized multi-subject text-to-video generation results by VideoDreamer. Given multiple subjects and few images for each subject, our VideoDreamer can generate videos that contain the given subjects, with new events and background, etc., guided by the text.}
    \label{fig:first page}
\end{figure*}

In this paper, we take a further step and investigate the more challenging task of customized multi-subject text-to-video generation. Given multiple user-defined subjects and few images for each subject, customized multi-subject text-to-video generation aims to generate videos that show the subjects and simultaneously conform to the textual prompts. As shown in Figure~\ref{fig:first page}, in customized multi-subject text-to-video generation, the user can create new actions of the multiple subjects, e.g.,  \textit{``surfing''}, and a new background for the videos, e.g., \textit{``in the sea''}. Despite the expected fascinating results, customized multi-subject text-to-video generation still remains a largely unexplored field. Moreover, generating multiple subjects often suffers the attribute binding problem (the visual features of subjects are mixed together and different subjects look similar), making the task more challenging.

To tackle the problems, we propose the 
novel VideoDreamer framework, which can generate text-guided multi-subject videos where the visual features of each given subject are well-preserved. VideoDreamer utilizes the pretrained text-to-image model, Stable Diffusion, with additional temporal modules to maintain temporal consistency, as the base video generator. Then the base generator is customized for multiple subjects with the proposed finetuning strategy. Particularly, to tackle the attribute binding problem, we propose a Disen-Mix finetuning strategy that guides the model to preserve the visual features of each subject with an auxiliary task to denoise the mixed images of the given subjects. To alleviate the influence of the artifacts of the mixed images, we finetune the mixed images with disentangled embeddings. Moreover, the Human-in-the-Loop Re-finetuning strategy is proposed to further enhance VideoDreamer performance. Additionally, we present a disentangled motion customization strategy to finetune the temporal modules so that we can generate videos with both customized subjects and motions. To evaluate the customized multi-subject text-to-video generation results, we propose the MultiStudioBench benchmark, which contains various subjects and textual prompts, with comprehensive metrics to evaluate the generated videos in subject fidelity, prompt fidelity, temporal consistency, etc. Our contributions are summarized as follows:

\begin{itemize}
    \item To the best of our knowledge, this work represents the first endeavor in the domain of customized multi-subject text-to-video generation.
    \item We propose a novel VideoDreamer framework, which customizes the text-to-video generator for multiple subjects by the proposed Disen-Mix Finetuning and Human-in-the-Loop Re-finetuning strategy, faithfully preserving the visual features of each subject.
    \item We present an effective disentangled motion finetuning strategy for VideoDreamer to support motion customization for the customized multiple subjects.
    \item We introduce MultiStudioBench, a benchmark tailored for evaluating customized multi-subject text-to-video generation models. Extensive experiments on MultiStudioBench demonstrate the remarkable generation capabilities of our proposed VideoDreamer.
\end{itemize}

\section{Related Work}
\label{sec:related work}
\paragraph{Text-to-image diffusion models} Diffusion models have shown a remarkable ability to learn data distributions, attracting attention from both academia and industry. Trained on large-scale text-image pairs, diffusion models~\cite{Imagen,dalle,glide,Sdiffusion,dalle2} can generate photo-realistic images based on the given textual prompts. GLIDE~\cite{glide} introduces classifier-free guidance to achieve better text control on images. Dall-E 2~\cite{dalle2} and Imagen~\cite{Imagen} utilize pretrained text models to further improve generation quality. Stable Diffusion (SD)~\cite{Sdiffusion} proposes to conduct diffusion process in the latent space, gaining speed and efficiency improvement while still maintaining a high resolution.

\paragraph{Text-to-video generation}  Driven by the success of text-to-image generation, the text-to-video task has received increasing attention recently. Text-to-video generation aims to generate temporal-coherent semantic videos that conform to the given textual prompts. Early works~\cite{earlyt2v1,earlyt2v2,earlyt2v3,earlyt2v4} primarily focus on simple-domain video generation, such as moving digits and human pose. Recently, pretrained on the large-scale video datasets~\cite{videodataset1,videodataset2,cogvideo}, both diffusion-based models~\cite{latentvideo,magicvideo,makeavideo,videofusion,free-bloom,text2video0,makeyourvideo,guo2023animatediff} and non-diffusion-based models~\cite{cogvideo,phenaki,godiva} are developed to generate more realistic and diverse videos. Despite the progress, the general text-to-video generation models cannot satisfy the personalized requirement for user-customized subjects.

\paragraph{Text-guided video editing}  Text-guided video editing aims to edit the content of the reference video with textual prompts~\cite{video-p2p,tune-a-video,vid2vid-0,controlvideo,dreamix,render-a-video,fatezero,text2video0}. Note that text-guided video editing is different from text-to-video generation, where the former requires an input video while the latter does not. Additionally, it is hard for text-guided video editing to change the motion or generate videos with new events.

\begin{figure}[htbp]
    \centering
    \includegraphics[width=\linewidth]{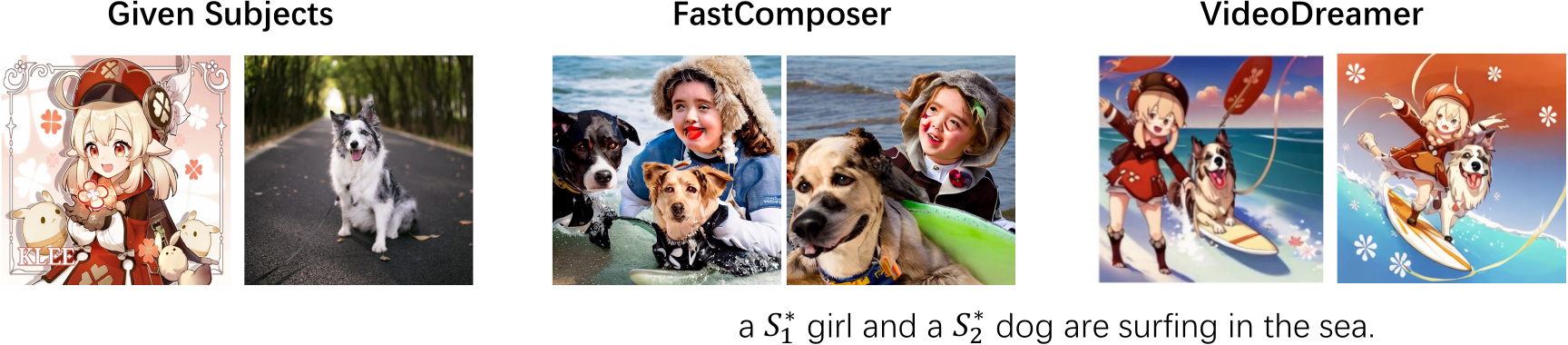}
    \caption{Visual comparison, where we use FastComposer and VideoDreamer to generate 2 images with 2 random seeds with the given prompt.}
    \label{fig:fastcomposer}
\end{figure}

\paragraph{Subject customization} 
Most subject customization works are still in the field of image generation. On one hand, some of the existing methods~\cite{dreambooth,TI,customfusion,svdiff,disenbooth} require finetuning on few images of the given subject, such as DreamBooth~\cite{dreambooth}, so that the subject can be reversed into a special text token. Consequently, customized generation can be achieved with the special token. Among the finetuning methods, \cite{dreambooth,TI,customfusion} face the attribute binding problem when applied to multiple subjects. \cite{svdiff} solves the attribute binding problem for multiple subjects by augmented data but introduces artificial stitches. \cite{mix-of-show} aims at a decentralized scenario for multiple subjects. On the other hand, other works~\cite{elite,Aslearning,Instantbooth,fastcomposer,subject-diffusion,break-a-scene} use additional datasets to train a module that can map an image to a text token for customization, making them free of the finetuning steps. Among the non-finetuning methods, \cite{elite,Aslearning,Instantbooth} are for single-subject customization, while \cite{fastcomposer,subject-diffusion} also consider the multi-subject scenario with attention controls for the attribute binding problem. However, these non-finetuning methods will fail to customize the subjects that are out-of-domain of the additional datasets, and therefore an effective finetuning strategy is still necessary. As shown in Fig.~\ref{fig:fastcomposer}, we use the non-finetuning method FastComposer~\cite{fastcomposer} to customize the cartoon girl and the dog, it will easily fail because the additional datasets it utilizes only contain real-world humans. The cartoon girl and dog are out-of-domain concepts for it. Additionally, in text-to-video generation, \cite{text2video0,free-bloom,guo2023animatediff} apply the image customization method DreamBooth to video models, and there are some other attempts~\cite{videoassembler,videobooth,dreamvideo} specifically designed for customized video generation. However, these methods are still limited to the single-subject scenario, failing to tackle the multi-subject video customization problem, whereas the static and dynamic attributes of multiple subjects are of significance to the visual big models~\cite{wang2024visual}.

\section{Method}
\label{sec:method}
\begin{figure*}[htbp]
    \centering
    \includegraphics[width=\linewidth]{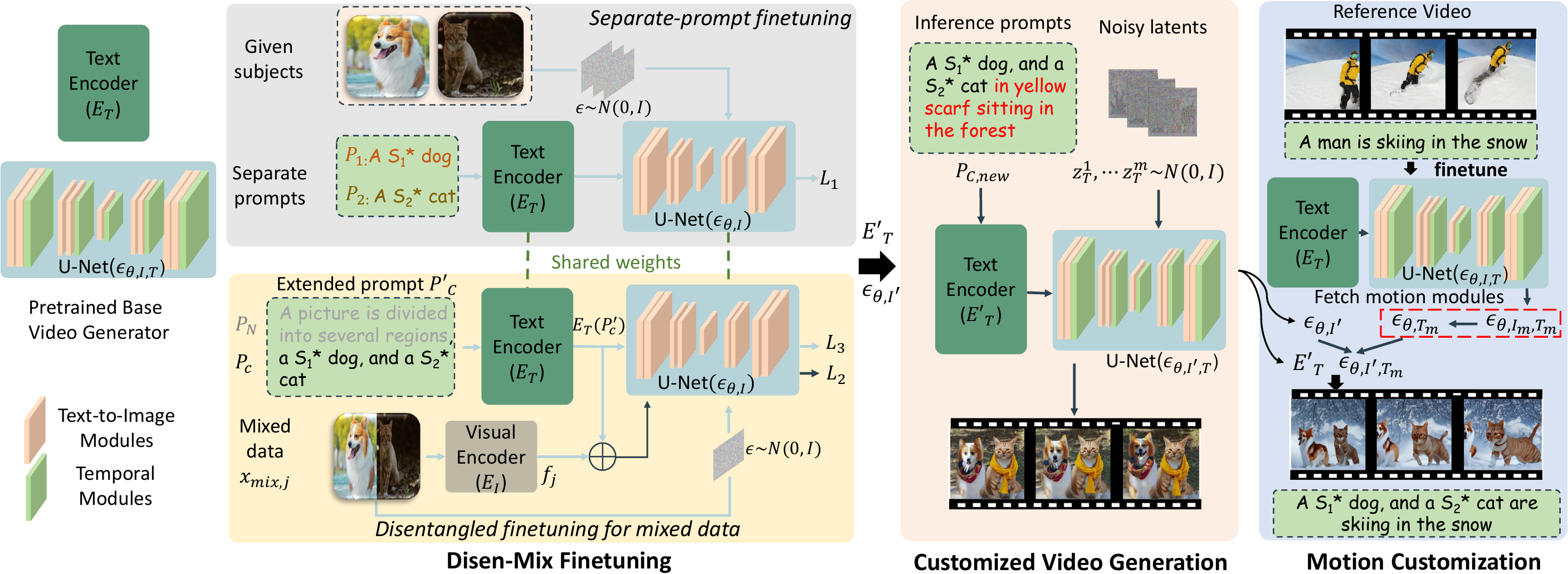}
    \caption{VideoDreamer: Given a pretrained video generator containing a text encoder $E_T$ and U-Net with motion modules $\epsilon_{\theta,I,T}$, in the Disen-Mix Finetuning, we finetune $E_T$ and the image modules $\epsilon_{\theta,I}$, where the separate-prompt finetuning is to customize each subject independently, while the disentangled finetuning for mixed data tackles the attribute binding problem. After finetuning, we obtain $E'_T$ and $\epsilon_{\theta,I'}$, which can be used to generate customized videos for multiple subjects. Additionally, we present a motion customization method, where we finetune the whole base text-to-video model on the reference video, and only use the finetuned motion modules $\epsilon_{\theta,T_m}$ together with image finetuned $E'_T$ and $\epsilon_{\theta,I'}$ to obtain videos with both customized motion and customized subjects.  }
    \label{fig:framework}
\end{figure*}

The overall VideoDreamer framework is shown in Figure~\ref{fig:framework}, which contains the Disen-Mix Finetuning stage for multi-subject customization, the customized video generation stage, and the motion customization stage. Next, we will introduce preliminaries about Stable Diffusion, present the base video generator, and our details about the VideoDreamer framework. 

\subsection{Preliminaries}
\label{subsec:pre}
\paragraph{Stable Diffusion} Stable Diffusion~\cite{Sdiffusion} is a pretrained text-to-image model on large-scale text-image pairs $\{ (P,x)\}$, where $x$ is an image and $P$ is the text description of the image $x$. To improve efficiency, Stable Diffusion conducts the forward and backward process in the latent space, with an encoder $\mathcal{E}(\cdot) $ and a decoder $\mathcal{D}(\cdot)$. The encoder transforms the image $x$ into the latent space, $z = \mathcal{E}(x)$, and the decoder reconstructs the image from the latent space with $x \approx \mathcal{D}(z)$, where $z$ is the latent code. Denoting the latent code of the image as $z_0$, next, we respectively describe the diffusion forward, backward, and training process. 

In the diffusion forward process, Gaussian noise is added to the latent code iteratively:
\begin{equation}
\small
q(z_t|z_{t-1}) = \mathcal{N}(z_t;\sqrt{1-\beta_t}z_{t-1},\beta_t I), t = 1,\cdots, T,
\end{equation}
where $T$ is a large step so that $z_T$ is close to a standard Gaussian noise.  

In the backward process (also called the sampling process), the Stable Diffusion will recover the image latent code $z_0$ from the noise $z_T$ step by step. Specifically, the denoising process relies on a U-Net~\cite{Unet}, which we denote as $\epsilon_{\theta,I}(\cdot)$, to predict the noise at each step. The U-Net is composed of convolutional and attentional (both self-attention and cross-attention) blocks. It receives the noisy latent code $z_t$, timestep $t$, and the textual feature $E_T(P)$ as input, and predicts the noise $\epsilon_{\theta,I}(z_t,t,E_T(P))$ at timestep $t$, where $E_T(\cdot)$ is a CLIP text encoder to encode the text prompt $P$. Then we get a less noisy latent code $z_{t'}$:
\begin{equation}
z_{t'} = Sampler(z_t, \epsilon_{\theta,I}(z_t,t,E_T(P));t',t) , t' < t, 
\end{equation}
where $Sampler(\cdot)$ could be DDPM~\cite{DDPM}, DDIM~\cite{DDIM}, or DPMSolver sampler~\cite{DPM}, and $t'$ also relies on the choice of the sampler, since different samplers require different sampling (backward) steps. The sampling process is conducted iteratively until we obtain $z_0$, and then we can map the latent code $z_0$ to the image $x$ with $x=\mathcal{D}(z_0)$.

To train the U-Net $\epsilon_{\theta,I}(\cdot)$, the following objective is usually adopted~\cite{DDPM,DDIM}:
\begin{equation}
\label{eq:sd obj}
\min~\mathbb{E}_{P,z_0,\epsilon,t}[ ||\epsilon-\epsilon_{\theta,I}(z_t, t, E_T(P))||_2^2  ],
\end{equation}
where for a randomly sampled noise $\epsilon$, we add it to the latent code $z_0$ and obtain the noisy latent $z_t$. What the U-Net $\epsilon_{\theta,I}(\cdot)$ needs to do is to make the predicted noise close to the sampled noise $\epsilon$. This objective will also be used during our finetuning for customization.

\subsection{Base Text-to-Video Generator} 
Inspired by~\cite{text2video0,guo2023animatediff}, we adopt the pretrained text-to-image Stable Diffusion model, equipped with temporal modules to maintain frame consistency, as the base text-to-video generator. On the one hand, the prior of Stable Diffusion can help to generate high-quality frames and diversified content. On the other hand, in such a generator, the text-to-image modules and temporal modules are decoupled, and it is natural to utilize images of the given multiple subjects to finetune the text-to-image modules, while fixing the temporal modules to preserve their ability to maintain frame consistency, which gives an elegant solution to the challenging customized multi-subject text-to-video generation task. Specifically, we choose two open-source pretrained text-to-video models, Text2video-Zero~\cite{text2video0}, and AnimateDiff~\cite{guo2023animatediff}. Assuming that we expect to generate a video of $m$ frames, we need first to prepare $m$ latent codes $\{z_T^{1},z_T^{2},\cdots,z_T^{m}\} \sim N(0,I)$ and send them to the Stable Diffusion model to denoise. However, directly denoising on the $m$ frames will result in $m$ independent frames, instead of a video. To tackle the problem, Text2video-Zero changes the self-attention in Stable Diffusion model to cross-frame attention to maintain frame consistency. AnimateDiff trains additional temporal modules on video datasets, which can be inserted into the Stable Diffusion model to generate videos. In our VideoDreamer framework, we try to customize Text2video-Zero and AnimateDiff with the given multiple subjects, where we can elegantly finetune the text-to-image Stable Diffusion modules with the images of the subjects. For simplicity, we denote the text-to-video generation process as:
\begin{equation}
\label{eq:t2v}
Vid = T2V(P;E_T,\epsilon_{\theta,I,T}),
\end{equation}
where $Vid$ is the output video, $T2V$ means the AnimateDiff or Text2video-Zero generator, $P$ is the prompt, $E_T$ is the text encoder, $\epsilon_{\theta,I,T}$ is the Stable diffusion with motion modules. Specifically, $\epsilon_{\theta,I,T}$ is composed of two decouples parts, i.e., the text-to-image modules $\epsilon_{\theta,I}$ and the motion modules $\epsilon_{\theta,T}$. Next, we will show how we finetune the parameters to achieve customized video generation. 

\subsection{Disen-Mix Finetuning}
\label{subsec:finetune}
Assume that there are $N$ user-defined subjects $\{ s_i\}_{i=1}^{N}$, and few images for each subject $\{ x_{ij} \}_{j=1}^{M_{i}} $, where $x_{ij}$ is the $j^{th}$ image of subject $s_i$ and $M_{i}$ (usually 3$\sim$5) is the number of images used for subject $s_i$. Disen-Mix Finetuning aims to provide the customized parameters $\epsilon_{\theta,I'}$ to generate videos for the given subjects, through finetuning the model on given images of the subject $ \{ \{ x_{ij} \}_{j=1}^{M_{i}} \}_{i=1}^N $. Specifically, Disen-Mix Finetuning contains separate-prompt finetuning for each subject, together with the disentangled finetuning for the mixed multi-subject data as follows.

\noindent \textbf{Separate-prompt finetuning} \ Similar to previous works~\cite{dreambooth,TI,disenbooth,customfusion}, we will first bind each subject $s_i$ to a special separated text prompt $P_i$,
where $P_i$ = ``a'' + ``$S_i^*$''+ ``$cate_i$'', and $cate_i$ is the category of subject $s_i$, such as \textit{``dog''},  and $S_i^*$ is a special token for the subject identity. The binding process is performed by finetuning the Stable Diffusion with a similar objective to Eq.~\ref{eq:sd obj} as follows:
\begin{equation}
\label{eq:separate obj}
\mathcal{L}_1 = \sum_{i=1}^N ( \sum_{j=1}^{M_i} \mathbb{E}_{\epsilon,t}[ ||\epsilon-\epsilon_{\theta,I}(z_{ij,t}, t, E_T(P_i))||_2^2  ]),
\end{equation}
where $z_{ij,t}$ is the noisy latent code of the $j^{th}$ image for subject $s_i$ at timestep $t$. The inner sum of the objective $\sum_{j=1}^{M_i} \mathbb{E}_{\epsilon,t}[ ||\epsilon-\epsilon_{\theta,I}(z_{ij,t}, t, E_T(P_i))||_2^2  ]$ means when we give the text prompt $P_i$, the model can denoise all the noisy latents for subject $s_i$, i.e., $\{ z_{ij,t}\}_{j=1}^{M_i}$ for all t, thus binding $P_i$ to the subject $s_i$. The outer sum means the same operation will be conducted for each subject, thus finishing the customization for all the given subjects.  

\begin{figure}[htbp]
    \centering
    \includegraphics[width=\linewidth]{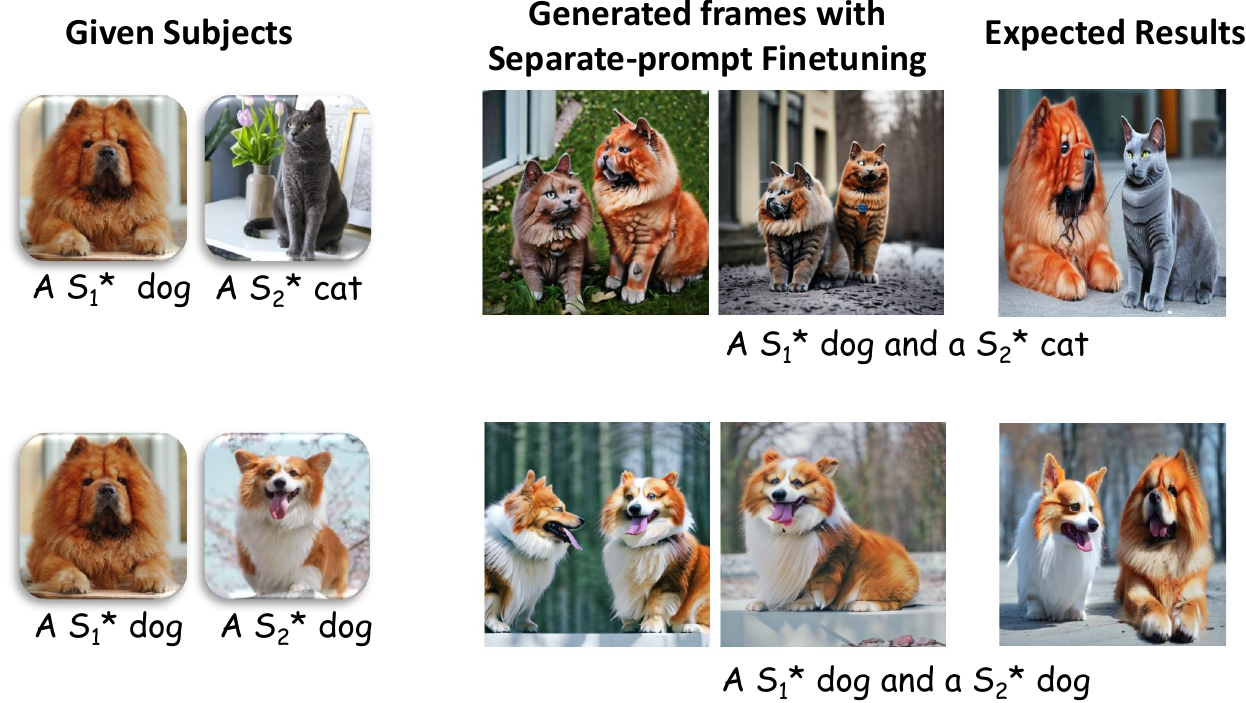}
    \caption{\label{fig:app:separate}Generated video frames only using separate-prompt finetuning, and the results are with 2 different random seeds. Only with the separate-prompt finetuning, the attributes of different subjects are mixed together. Sometimes one subject is missing.}
\end{figure}

Now, it is natural to directly use the concatenation of all the prompts $P_c = [P_1, P_2, \cdots, P_N]$ (e.g., \textit{``a $S_1^*$ dog, a $S_2^*$ cat''}) and the finetuned parameters to generate videos of all the given subjects. However, this naive strategy will face the attribute binding and object missing problem as shown in Figure~\ref{fig:app:separate}, motivating us to propose the following disentangled finetuning for the mixed multi-subject data.   

\noindent \textbf{Disentangled finetuning for mixed multi-subject data} The reason why the model with separate finetuning fails to simultaneously customize multiple subjects is that $P_c$ is a new token to the model, which is not seen by the model during finetuning. Relying on the prior of the Stable Diffusion to compose the separately-finetuned subjects into one image will inherit its attribute binding and missing object problem~\cite{attr,customfusion}. To provide further guidance to multi-subject generation, we mix the images of different subjects into one image as follows,
\begin{equation}
x_{mix} = [x_{1m_1};x_{2m_2};\cdots;x_{Nm_N} ] ,
\end{equation}
where $x_{im_i}$ is a randomly sampled image for subject $s_i$, and $[ ; ]$ is the concatenation operation. By sampling different images for each subject, we can obtain different mixed images. Consequently, we obtain a small dataset $D_{mix} = \{ x_{mix,j}\}_{j=1}^{M_{mix}}$ of $M_{mix}$ images, where each image contains all the given subjects, which can be bind to the mixed prompt $P_c$.  However, simply binding $P_c$ with the $D_{mix}$ with the previous finetuning strategy will make the generated images using $P_c$ suffer from artifacts, e.g., the generated images will contain stitches introduced by the concatenation. To alleviate the influence of artifacts, we propose a finetuning strategy with disentangled embeddings inspired by the single-subject customization work~\cite{disenbooth}. 

Instead of directly using $P_c$ as the condition to denoise the images in $D_{mix}$, we introduce the disentangled image-specific condition, shared stitch condition, and shared subject-identity condition $P_c$ together to denoise. The idea behind the design is that each image in $D_{mix}$ not only contains multiple subjects, but also artificial stitches, and image-specific information such as the background and the subject pose. To describe each image, we first extend $P_c$ with a stitch prompt $P_N=$\textit{``a picture is divided into several regions''} and obtain $P'_c = [P_N, P_c]$, e.g., \textit{``a picture is divided into several regions, a $S_1^*$ boy, a $S_2^*$ dog, and a $S_3^*$ cat''}. Then, we can obtain the textual condition embedding $E_T(P'_c)$ through the CLIP text encoder $E_T(\cdot)$. To further obtain the image-specific embedding, we use a CLIP visual encoder followed by an adapter as follows,
\begin{equation}
    f_j = Adapter(E_I(x_{mix,j}) ), j=1,\cdots,M_{mix},
\end{equation}
where $E_I(\cdot)$ is the pretrained CLIP visual encoder, and $Adapter(\cdot)$ is an MLP adapter with skip connection. With the textual and image-specific embedding, we can denoise the images of mixed data as follows,
\begin{equation}
\label{eq:precise}
\mathcal{L}_2 = \sum_{j=1}^{M_{mix}} \mathbb{E}_{\epsilon,t}[ ||\epsilon-\epsilon_{\theta,I}(z_{mix,j,t}, t, E_T(P'_c)+f_j)||_2^2  ],
\end{equation}
where for the noisy latent code $z_{mix,j,t}$ of each mixed image at timestep $t$, we use the sum of the textual embedding and the visual embedding to denoise it. With the extended prompt and the visual embedding, $P_c$ can focus on the information about the subjects that it describes, while letting the stitch prompt $P_N$ and the visual embedding $f_j$ denoise the stitches and subject-irrelevant information. Considering that $f_j$ is an image-specific feature that may capture all information of image $x_{mix,j}$, causing $E_T(P'_c)$ to contain insufficient subject information, to avoid this problem, we adopt the weak denoising objective as~\cite{disenbooth}:
\begin{equation}
\label{eq:weak}
\mathcal{L}_3 = \lambda\sum_{j=1}^{M_{mix}} \mathbb{E}_{\epsilon,t}[ ||\epsilon-\epsilon_{\theta,I}(z_{mix,j,t}, t, E_T(P'_c))||_2^2  ],
\end{equation}
where $\lambda < 1$ is a hyper-parameter set to 0.01 as given in~\cite{disenbooth}. The weak denoising objective plays as a regularizer to make $E_T(P'_c)$ denoise the mixed image, preventing it from losing subject visual details, but $\lambda$ should not be too large, or $E_T(P'_c)$ may overfit the subject-irrelevant information.

In sum, finetuning the Stable Diffusion model on the following objective, $P_c$ can be used as the prompt for the given multiple subjects while being not influenced by the artifacts.
\begin{equation}
\label{eq:total}
\mathcal{L} = \mathcal{L}_1 + \mathcal{L}_2 + \mathcal{L}_3 .
\end{equation}

\subsection{Optional: Human-in-the-Loop Re-finetuning}
\label{subsec:human}
To further improve the multi-subject generation performance, we present the Human-in-the-Loop Re-finetuning strategy(HLR). Specifically, we will first use the Disen-Mix Finetuning to obtain a finetuned Stable Diffusion model $\epsilon_{\theta,I_1}$, and then use $P_c$ and some related prompts, e.g., $P_c$ + \textit{``in the ocean''} or \textit{``in the flowers''}, to generate some pictures about the given multiple subjects. Then, we can pick few satisfying pictures from the generated pictures by humans. After that, we can re-finetune the Stable diffusion model using Eq.~\ref{eq:total} by replacing the original mixed images with the picked images. Note that here we change $P'_c$ in Eq.~\ref{eq:weak} and Eq.~\ref{eq:precise} to $P_c$, because there are no stitches in these picked images and we do not need the extended prompt \textit{``a picture is divided into several regions''} anymore. The re-finetuned model will bring better performance for some hard cases. In our main experiments, we do not apply HLR for comparison, but we conduct an ablation about its effectiveness.

\subsection{Parameters to Finetune and Inference}
The parameters to finetune contain the mentioned adapter. Additionally, we apply LoRA~\cite{LoRA} to finetune the U-Net and text encoder. The finetuned text encoder and U-Net are denoted as $E'_T$ and $\epsilon_{\theta,I'}$. To generate videos about the customized multiple subjects, we combine $P_c$ with many other prompts, e.g., \textit{``surfing in the ocean''}, to obtain $P_{c,new}$. Finally, we can generate new videos using Eq.~\ref{eq:t2v} as $Vid = T2V(P_{c,new};E'_T,\epsilon_{\theta,I',T})$, where we use the finetuned image modules, text encoder, $P_{c,new}$ together with the temporal modules to generate customized videos for multiple subjects.

\subsection{Motion Customization}
Besides customizing the given multiple subjects, we also present a disentangled finetuning strategy for motion customization, which enables users to generate videos of both customized subjects and motions with VideoDreamer. Specifically, given a reference video $V_m$, and its text prompt $P_m$ (e.g., \textit{``a man is skiing in the snow''}). We use the text-video pair to finetune the base text-to-video generator and we will obtain the finetuned model $\epsilon_{\theta,I_m,T_m}$, where the image modules and motion modules are all finetuned. Then, it is a natural idea to apply previous LoRA parameters to the $\epsilon_{\theta,I_m,T_m}$ for both subject and motion customization (we call this method Naive-motion in the experiments). However, we find that $\epsilon_{\theta,I_m,T_m}$ easily overfits the appearance of the reference video and it is hard to generate the customized subjects. Therefore, inspired by the idea of image-motion disentanglement, as shown in Figure~\ref{fig:framework}, we abandon the image modules $\epsilon_{\theta,I_m}$ that mainly involve the subject appearance from $\epsilon_{\theta,I_m,T_m}$, and only use the motion modules $\epsilon_{\theta,T_m}$, and combine it with $\epsilon_{\theta,I'}$ to obtain $\epsilon_{\theta,I',T_m}$. Finally, with $E'_T$, $\epsilon_{\theta,I',T_m}$, and prompt $P_{c,m}$(\textit{``a $S_1^*$ dog, a $S_2^*$ cat'' are skiing in the snow}''), we can achieve both subject and motion customization.

\section{Experiments}
\label{sec:experiment}

\subsection{Experimental Settings}
\begin{figure}
    \centering
    \includegraphics[width=0.7\linewidth]{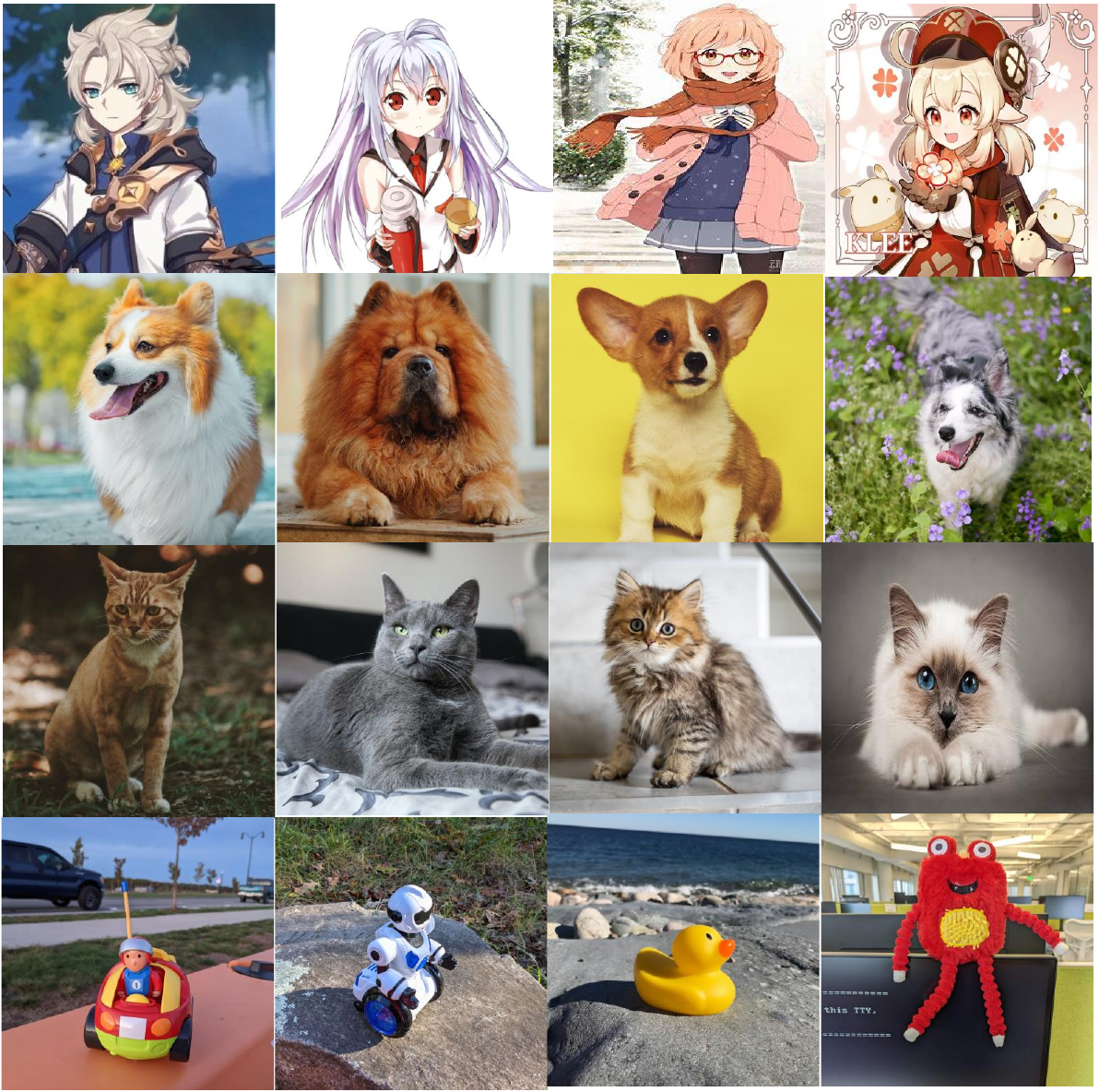}
    \caption{Part of the MultiStudioBench dataset images.}
    \label{fig:data}
\end{figure}

\begin{figure}
    \centering
    \includegraphics[width=0.9\linewidth]{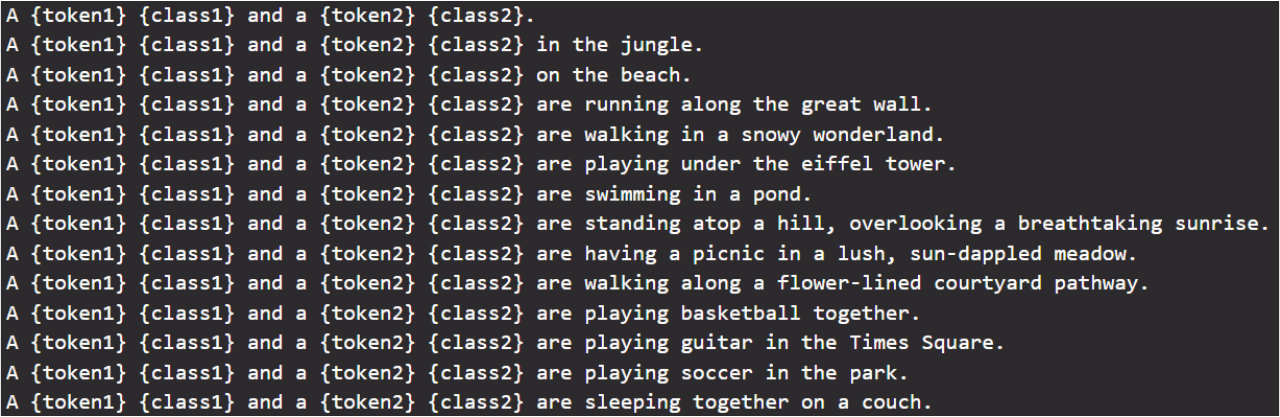}
    \caption{Part of the evaluation prompts for two-subject combinations.}
    \label{fig:data2}
\end{figure}

\noindent\textbf{Dataset.} \ Since this is the first work for customized multi-subject text-to-video generation, we propose the MultiStudioBench dataset. The dataset contains 25 subjects, including personal belongings, pets, and some animation characters, and there are few images for each subject. Images in the dataset are from previous works~\cite{dreambooth,customfusion} or collected by the authors. We provide an overview of part of the datasets in Figure~\ref{fig:data}, where we can see that the images are very diverse, covering different categories and styles. Among the collected subjects, we selected 15 combinations for customization in total, including 12 2-subject combinations (e.g., a cat and a dog) and 3 3-subject combinations (e.g., a cat, a dog, and a toy). We also provide 30 textual prompts used for the generation, where the textual prompts are designed to generate new actions of subjects (e.g., \textit{``playing chess, sleeping''}), new backgrounds (e.g., \textit{``under the Eiffel tower''}), etc. We provide part of the evaluation prompts for two-subject combinations in Figure~\ref{fig:data2}. For a more robust evaluation, we generate videos with 4 random seeds for each subject combination and each prompt, totaling 1800 videos.

\noindent\textbf{Baselines.} \ There is no existing work for customized multi-subject text-to-video generation to directly compare with. However, considering that our work is built on finetuning the base video generator (Text2video-zero/AnimateDiff) in a customized way, we can replace the Disen-Mix finetuning strategy in VideoDreamer with some customized finetuning
strategies. Specifically, we adopt the DreamBooth\cite{dreambooth}, Customfuison\cite{customfusion}, and SVDiff\cite{svdiff} for customization, respectively obtain the DB+AD/T2V, Custom+AD/T2V and SVDiff+AD/T2V baselines, where AD and T2V are short for AnimateDiff and Text2video-Zero, respectively.

\noindent \textbf{Implementation Details.}\ Our code is based on Diffusers~\cite{hugging}, where we use the pretrained Stable Diffusion 2-1 for Text2video-Zero, and pretrained Stable Diffusion 1-5 for AnimateDiff. During finetuning, we adopt the AdamW~\cite{adamw} optimizer, with the text encoder learning rate $1e-5$. The learning rate for other parameters is $5e-5$ for 2-subject customization and $1e-4$ for 3-subject customization. During inference, the video is 8-frame for Text2video-Zero and 16-frame for AnimateDiff, with resolution 512 $\times$ 512. For Text2video-Zero, we adopt the DPMSolver as the video-generator sampler, where we set $T=40, T'=38$, while other hyper-parameters as default in~\cite{hugging}. For AnimateDiff, we adopt the default scheduler and hyper-parameters for inference as~\cite{hugging}.

\noindent\textbf{Metrics.} \ MultiStudioBench evaluates the generated videos from 4 aspects. (i) Subject fidelity: generated videos should contain the given customized subjects. For the frames in the generated video, we first use the pretrained detection model FasterRCNN-MobileNet-V3-large~\cite{pytorch} to detect the subjects, and calculate the \textbf{DINO} score between the detected subjects and the given subjects, where the DINO score is the DINO image feature cosine similarity proposed by~\cite{dreambooth}. (ii) Textual fidelity: the generated videos should be consistent with the given textual prompt. We use the average \textbf{CLIP-T} score~\cite{dreambooth,TI} between each frame and the given textual prompt to evaluate the textual fidelity of the generated video. (iii) \textbf{Temporal Consistency}: We use the average CLIP image cosine similarity between all pairs of video frames to measure the temporal consistency of the video as in~\cite{tune-a-video}. (iv) \textbf{Stitch Score}: This metric is used to distinguish the methods like SVDiff that may introduce the artificial stitches. We use OpenCV~\cite{opencv} tools to detect whether each frame has artificial stitches. If the stitches are detected in the frame, the stitch score of the frame is 1.0, or the score is 0.0. We finally report the average stitch score on all the frames. A lower stitch score indicates better performance. 

\subsection{Main Results}
\begin{figure*}[htbp]
    \centering
    \includegraphics[width=0.8\linewidth]{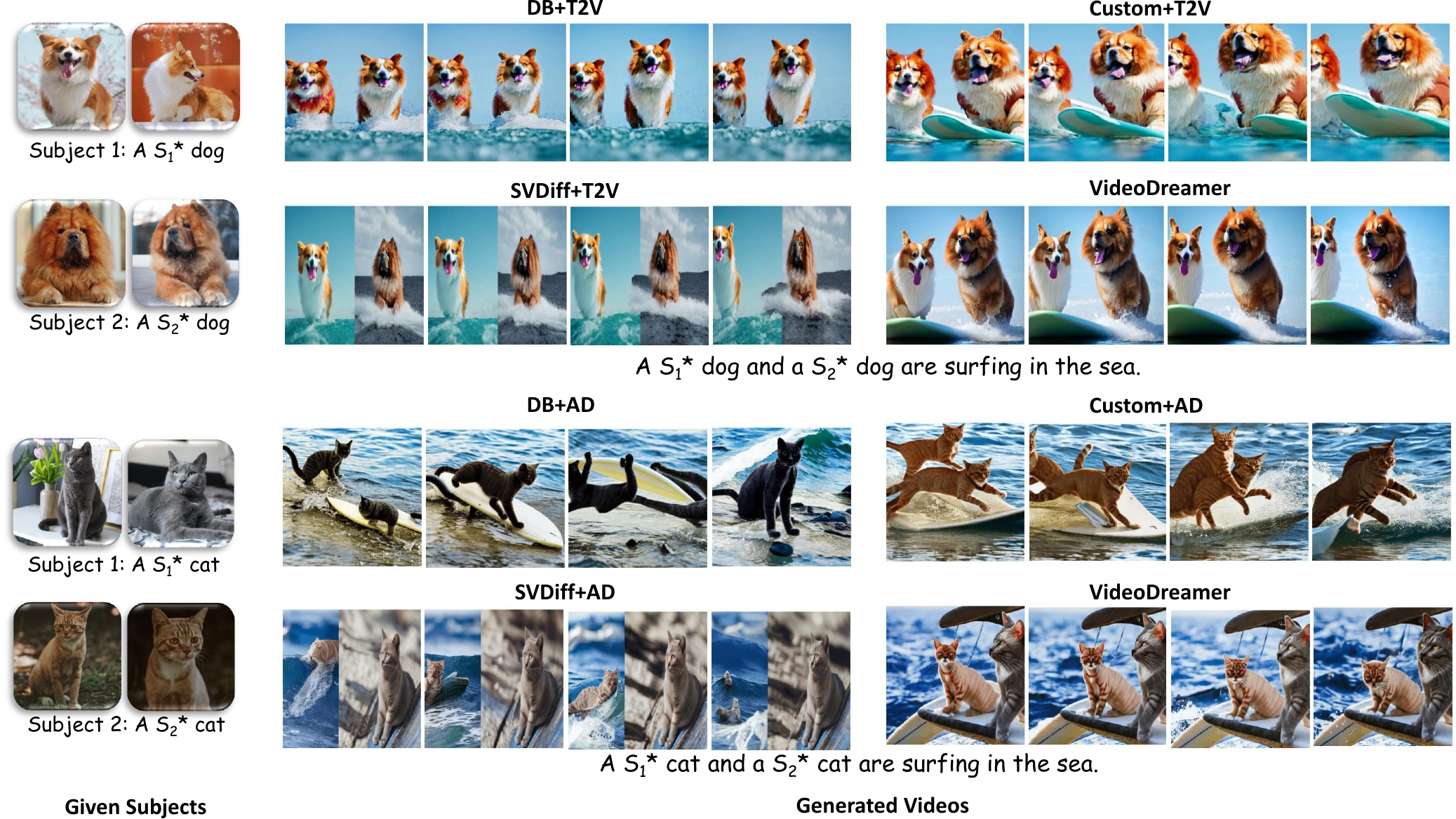}
    \caption{Qualitative comparison between VideoDreamer and baselines. Baselines suffer from attribute binding, missing subjects problems or artifacts. VideoDreamer can faithfully generate videos that contain the given 
    subjects and conform to the textual prompts. }
    \label{fig:compare}
\end{figure*}

\paragraph{Qualitative results} The qualitative results are presented in Figure~\ref{fig:compare}. We can see DB and Custom suffer from attribute binding problems, e.g., the generated two subjects look similar. Additionally, when the base model is AnimateDiff, some subjects are missing, which causes low temporal consistency. SVDiff suffers from artifacts. In contrast, our VideoDreamer can generate temporally consistent videos that faithfully preserve the subject identity while alleviating the impact of artifacts.
\begin{table*}[t]
\small
\centering
\caption{Quantitative Comparison between VideoDreamer and baselines. 2-sbj and 3-sbj respectively indicate the average performance on 2-subject customization and 3-subject customization. Avg. indicates the average performance on all the data. The best average performance is in bold and second is underlined. $\uparrow$ indicates higher metric value represents better performance and vice versa.}
\label{tab:main}
\resizebox{\linewidth}{!}
{
\begin{tabular}{lcccccccccccc}
\toprule
                 & \multicolumn{3}{c}{\textbf{DINO}$\uparrow$} & \multicolumn{3}{c}{\textbf{CLIP-T}$\uparrow$} & \multicolumn{3}{c}{\textbf{Temporal Consistency}$\uparrow$} & \multicolumn{3}{c}{\textbf{Stitch Score}$\downarrow$} \\
\cmidrule(r){2-4} \cmidrule(r){5-7} \cmidrule(r){8-10} \cmidrule(r){11-13}
 &\multicolumn{1}{c}{2-sbj} &
  \multicolumn{1}{c}{3-sbj} &
  \multicolumn{1}{c}{\textbf{Avg.}} &
  \multicolumn{1}{c}{2-sbj} &
  \multicolumn{1}{c}{3-sbj} &
  \multicolumn{1}{c}{\textbf{Avg.}} &
  \multicolumn{1}{c}{2-sbj} &
  \multicolumn{1}{c}{3-sbj} &
  \multicolumn{1}{c}{\textbf{Avg.}} &
  \multicolumn{1}{c}{2-sbj} &
  \multicolumn{1}{c}{3-sbj} &
  \multicolumn{1}{c}{\textbf{Avg.}} \\
DB+AD & 0.362 & 0.315 & \underline{0.353} & 0.299 & 0.305 & 0.300 & 0.913 & 0.919 & 0.914 & 0.074 & 0.084 & \textbf{0.076} \\
Custom+AD & 0.359 & 0.310 & 0.349 & 0.308 & 0.309 & \textbf{0.309} & 0.896 & 0.905 & 0.898 & 0.125 & 0.187 & 0.137 \\
SVDiff+AD & 0.377 & 0.260 & \underline{0.353} & 0.281 & 0.296 & 0.284 & 0.928 & 0.954 & \underline{0.934} & 0.631 & 0.685 & 0.642 \\
VideoDreamer(AD) & 0.408 & 0.335 & \textbf{0.394} & 0.300 & 0.303 & \underline{0.301} & 0.937 & 0.933 & \textbf{0.936} & 0.107 & 0.183 & \underline{0.122} \\
\midrule
DB+T2V & 0.454 & 0.408 & 0.445 & 0.315 & 0.317 & \textbf{0.316} & 0.941 & 0.941 & 0.941 & 0.032 & 0.037 & \textbf{0.033} \\
Custom+T2V & 0.461 & 0.401 & \underline{0.449} & 0.314 & 0.309 & \underline{0.313} & 0.945 & 0.948 & \textbf{0.946} & 0.111 & 0.133 & 0.115 \\
SVDiff+T2V & 0.461 & 0.305 & 0.430 & 0.285 & 0.285 & 0.285 & 0.942 & 0.926 & 0.939 & 0.237 & 0.768 & 0.343 \\
VideoDreamer(T2V) & 0.493 & 0.428 & \textbf{0.480} & 0.313 & 0.314 & \underline{0.313} & 0.946 & 0.939 & \underline{0.944} & 0.034 & 0.178 & \underline{0.062}\\
\bottomrule
\end{tabular}
}
\end{table*}

\paragraph{Quantitative results} The overall quantitative results are reported in Table~\ref{tab:main}. From the results, we can observe that: (i) VideoDreamer achieves a much higher DINO score than all the baselines, indicating that it has the best subject fidelity and the best customization ability. (ii) Although VideoDreamer and SVDiff are finetuned on the mixed data, SVDiff suffers from overfitting the mixed data, thus having a low CLIP-T score on both AD and T2V base models. In contrast, the disentangled tuning strategy avoids VideoDreamer overfitting the identity-irrelevant information in the mixed data, achieving comparable text fidelity, i.e., CLIP-T score, to DB and Custom. (iii) The temporal consistency of all the methods on T2V base model is similar, while on the AD base model, VideoDreamer and SVDiff achieve clearly better temporal consistency than other methods, indicating their ability to stably customize multiple subjects in each frame, thus better maintaining temporal consistency. (iv) SVDiff has the highest stitch score and suffers from artifacts. Custom also has a high stitch score because it applies image-crop augmentation during finetuning, which introduces stitches. Our VideoDreamer and DB have a low stitch score, indicating the effectiveness of our Disen-Mix finetuning strategy. In sum, our proposed VideoDreamer has the best ability for customization, while also keeping a high textual fidelity, temporal consistency, and fewer artifacts.\\

\subsection{Ablation Study} 
\begin{figure*}[htbp]
    \centering
    \includegraphics[width=0.8\linewidth]{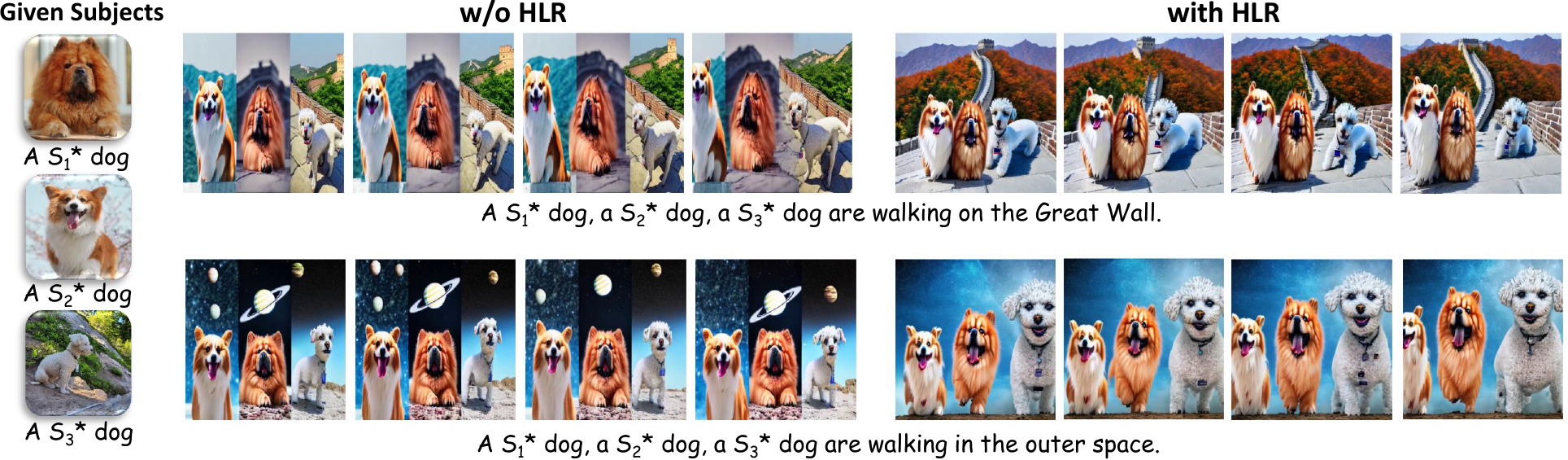}
    \caption{Qualitative results when VideoDreamer with and without HLR to customize 3 subjects.}
    \label{fig:HLR}
\end{figure*}

\begin{figure*}[htbp]
    \centering
    \includegraphics[width=0.8\linewidth]{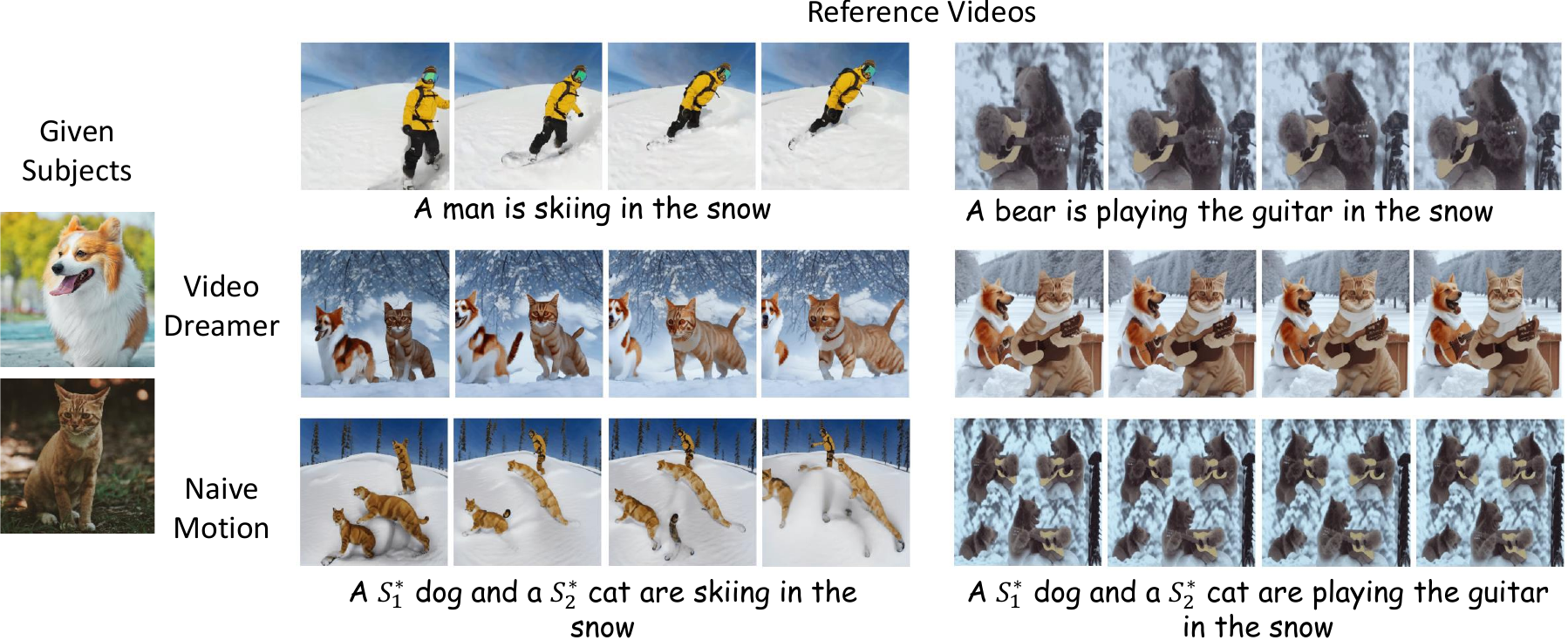}
    \caption{Joint subject and motion customization results on AD base model.}
    \label{fig:motion}
\end{figure*}

\begin{table}[htbp]
\centering
\caption{The effectiveness of the proposed Human-in-the-Loop Re-finetuning strategy(HLR) on the 3-subject scenario, where the base model is T2V. Temporal Consistency and Stitch Score are abbreviated as Temp Consist and Stit Score.}
\label{tab:hlr}
\resizebox{\linewidth}{!}{
\begin{tabular}{lcc}
\toprule
& VideoDreamer & VideoDreamer+HLR \\
\hline
\textbf{DINO}$\uparrow$       & 0.428       & \textbf{0.440}  \\
\textbf{CLIP-T}$\uparrow$     & 0.314       & \textbf{0.324}   \\
\textbf{Temp Consist}$\uparrow$ & \textbf{0.939}       & 0.937 \\
\textbf{Stit Score}$\downarrow$      & 0.178       & \textbf{0.045} \\            
\bottomrule
\end{tabular}
}
\end{table}
\paragraph{Human-in-the-Loop Re-finetuning} As shown in Table~\ref{tab:main}, the stitch score of VideoDreamer will increase when facing 3 subjects. To tackle this problem, we use the aforementioned Human-in-the-Loop Re-finetuning(HLR). The quantitative results are given in Table~\ref{tab:hlr}, and we can see that the proposed HLR largely reduces the impact of the stitches. The corresponding qualitative comparisons are given in Figure~\ref{fig:HLR}, further demonstrating the effectiveness of the HLR.

\begin{figure*}[h]
    \centering
    \includegraphics[width=0.9\linewidth]{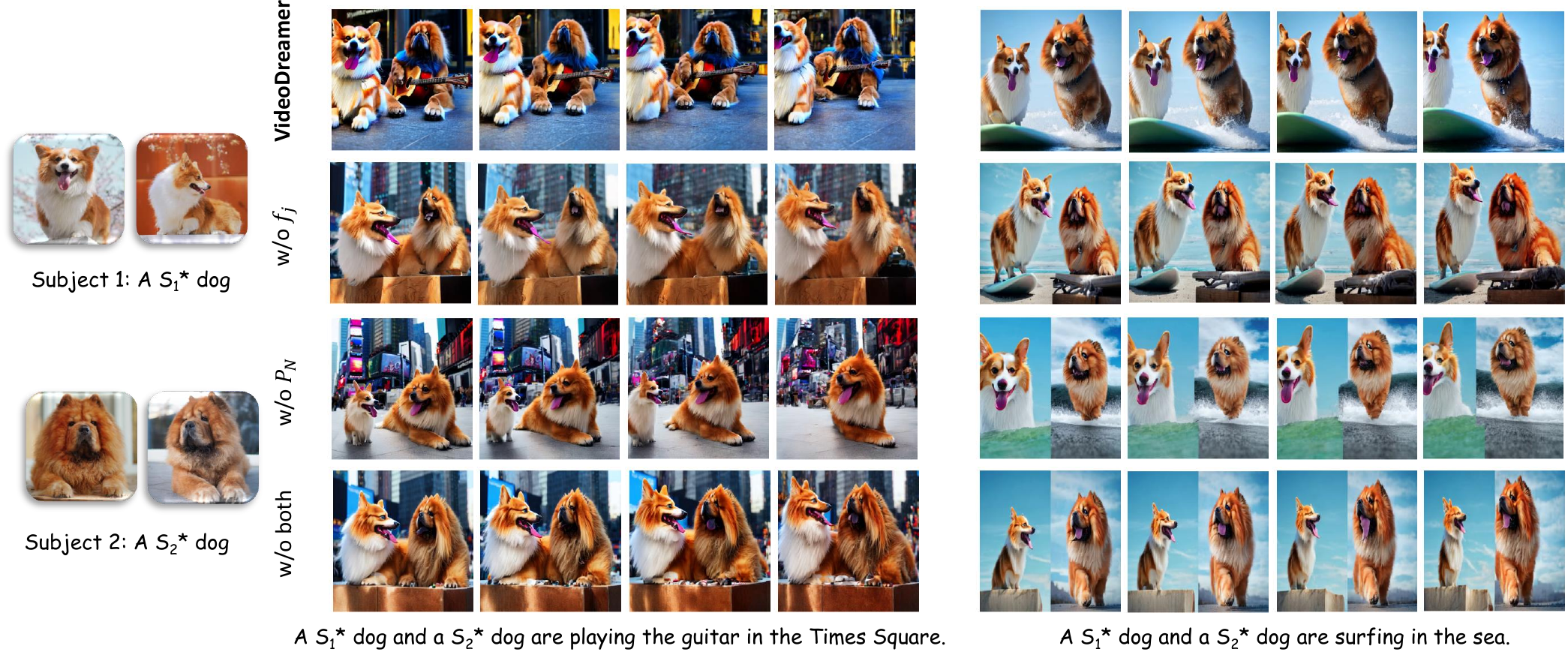}
    \caption{Qualitative results about the disentangled embedding ablation study.}
    \label{fig:ablaion-disen}
\end{figure*}

\paragraph{Disentangled embedding ablation} In VideoDreamer finetuning, besides the shared subject-identity condition $P_c$, we also use the shared stitch condition $P_N=$\textit{``a picture is divided into several regions''}, and the image-specific embedding $f_j$, to avoid overfitting the subject-irrelevant information. We validate their effectiveness in Table~\ref{tab:disen} on the T2V base model, where we randomly choose 4 2-subject combinations and report the average performance. From the results, we can see that both $P_N$ and $f_j$ can help to reduce the artificial stitches. Additionally, using the image-specific feature $f_j$ can prevent the model from overfitting the given images, improving the textual fidelity(CLIP-T), which is consistent with the results in~\cite{disenbooth}. Corresponding qualitative comparisons are presented in Fig.\ref{fig:ablaion-disen}. In the first example, from the results of w/o $P_N$ in \textit{``surfing in the sea''}, we can see that without the stitch prompt $P_N$, the generated videos may contain artificial stitches, showing the effectiveness of the stitch prompt to remove the artifacts. From the results of w/o $f_j$, we can see that without $f_j$, the generated videos may overfit some subject-irrelevant information, e.g., the stage of the given image in subject 1, and ignore the textual prompt, e.g., \textit{``playing the guitar''}. Therefore, the disentangled embeddings during training help to alleviate the impact of artifacts and improve the textual fidelity,

\paragraph{Weak denoising loss}To optimize the model, we introduce the weak denoising loss $L_3$, we present the ablation about it in Table~\ref{tab:app:loss} on the four subject combinations on Text2video-Zero as previous ablations. Without $L_3$(w/o $L_3$) to keep $P'_c$ containing mixed data information, the DINO score will decrease, which means the generated subject will be less similar to the given subjects, which is consistent with the results in~\cite{disenbooth}.

\begin{table}[htbp]
\caption{Ablations about the weak denoising loss.} 
\label{tab:app:loss}
\small
\centering
\begin{tabular}{lcccc}
\toprule
 & \multicolumn{1}{l}{\textbf{DINO}} & \multicolumn{1}{l}{\textbf{CLIP-T}} & \multicolumn{1}{l}{\textbf{Temp Consist}} & \multicolumn{1}{l}{\textbf{Stit Score}} \\
\hline
ours & \textbf{0.501} & \textbf{0.306} &0.943 &0.036 \\
w/o $L_3$     & 0.493 & 0.305 & \textbf{0.944} & \textbf{0.032} \\
\bottomrule
\end{tabular}
\end{table}

\paragraph{Joint subject and motion customization} We provide the motion customization results for multiple subjects in Figure~\ref{fig:motion}. The results show that our proposed motion customization method can preserve the appearance of each subject and inherit the motion of the reference video, but the naive baseline overfits the appearance of the reference videos.

\begin{table}[htbp]
\caption{Ablations about the disentangled embeddings.} 
\label{tab:disen}
\centering
\resizebox{\linewidth}{!}
{
\begin{tabular}{lcccc}
\toprule
 & \multicolumn{1}{l}{\textbf{DINO}$\uparrow$} & \multicolumn{1}{l}{\textbf{CLIP-T}$\uparrow$} & \multicolumn{1}{l}{\textbf{Temp Consist}$\uparrow$} & \multicolumn{1}{l}{\textbf{Stit Score}$\downarrow$} \\
\hline
ours & 0.501 & \textbf{0.306} & \textbf{0.943} & \textbf{0.036} \\
w/o $f_j$      & 0.486 & 0.299 & 0.936 & 0.125 \\
w/o $P_N$     & \textbf{0.504} & 0.305 & 0.942 & 0.068 \\
w/o both    & 0.476 & 0.296 & 0.938 & 0.287 \\
\bottomrule
\end{tabular}
}
\end{table}

\subsection{More results}
\paragraph{Evaluation on More Comprehensive Metrics} We also use the motion\_smoothness(abbreviated as motion), aesthetic\_quaility(aesthetic), and imaging\_quality(imaging), 3 metrics from~\cite{vbench} to evaluate different methods more comprehensively, where motion\_smoothness evaluates whether the generated video has a smooth motion, aesthetic\_quality and imaging\_quality evaluates the image quality of the video frames, larger values on the 3 metrics mean better performance. Additionally, we use human assessment to evaluate the quality of the generated videos. Specifically, we asked 50 users of different occupations to rank the videos generated by different methods, by jointly considering whether the generated videos have the same subjects as the given images, whether they are consistent with the text prompts and whether the video is temporally consistent and natural. For each user, we randomly sample 10 unique prompts, and we report the average rank, a smaller rank value(closer to 1) indicates better performance. The performance of different methods on these 4 new metrics is shown in Table~\ref{tab:more metrics}. The results further show that our proposed VideoDreamer has superior generation ability than existing finetuning methods.

\begin{table}[!ht]
    \caption{Evaluating different methods on more metrics.}
    \label{tab:more metrics}
    \centering
    \begin{tabular}{lcccc}
    \hline
        ~ & DB+AD & Custom+AD & SVDiff+AD & VideoDreamer \\ \hline
        motion & 0.928 & 0.901 & 0.924 & \textbf{0.943} \\ 
        aesthetic & 0.540 & 0.572 & 0.455 & \textbf{0.588} \\ 
        imaging & 0.728 & 0.726 & 0.726 & \textbf{0.735} \\ 
        Human & 2.966 & 2.415 & 3.082 & \textbf{1.537} \\ \hline
    \end{tabular}
\end{table}

\begin{figure}[htbp]
    \centering
    \includegraphics[width=\linewidth]{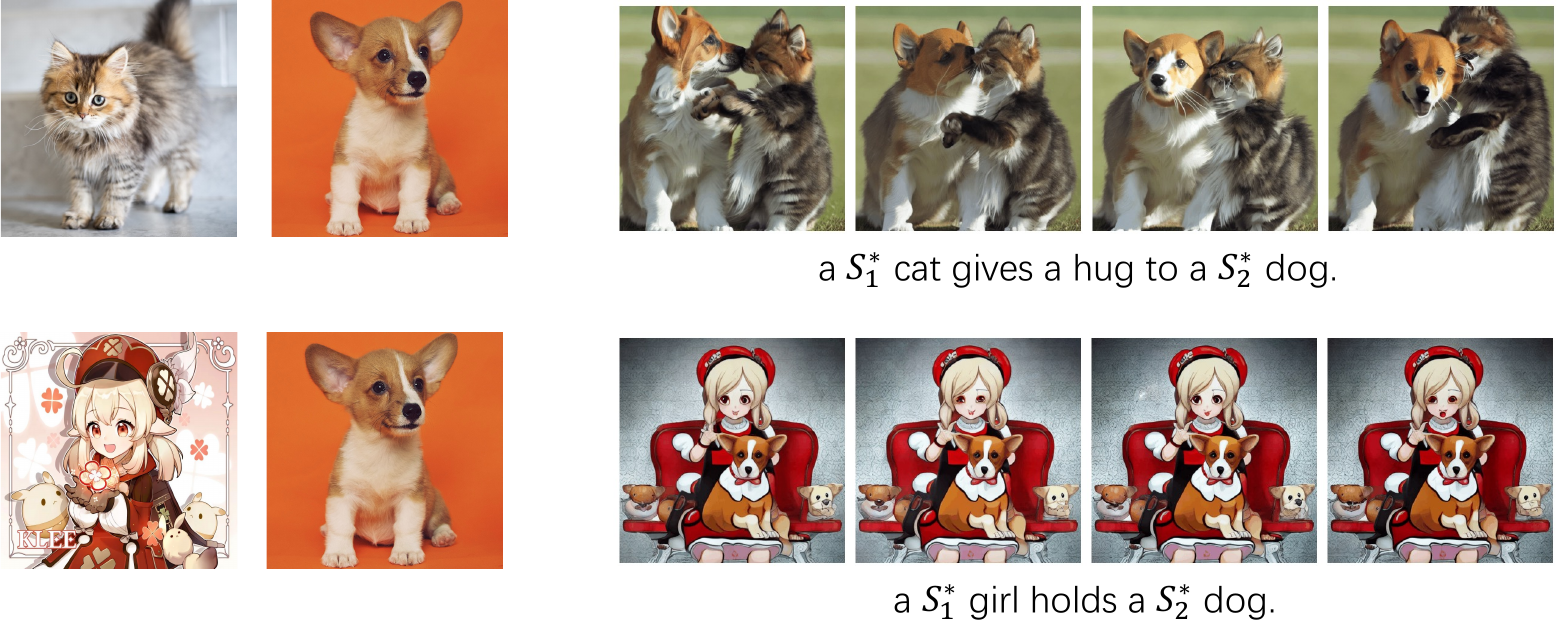}
    \caption{Generating subjects with more interactions.}
    \label{fig:interaction}
\end{figure}

\paragraph{Subject Interaction Generation} Since we stitch the resized images to an image that contains multiple subjects as guidance for multi-subject generation, we want to explore whether VideoDreamer can still generate images where the subjects have other interactions instead of in different regions. As shown in Figure~\ref{fig:interaction}, thanks to the disentangled finetuning strategy, our method does not overfit the stitched images and can generate interactions such as ``hold'' and ``gives a hug''.

\begin{figure}
    \centering
    \includegraphics[width=\linewidth]{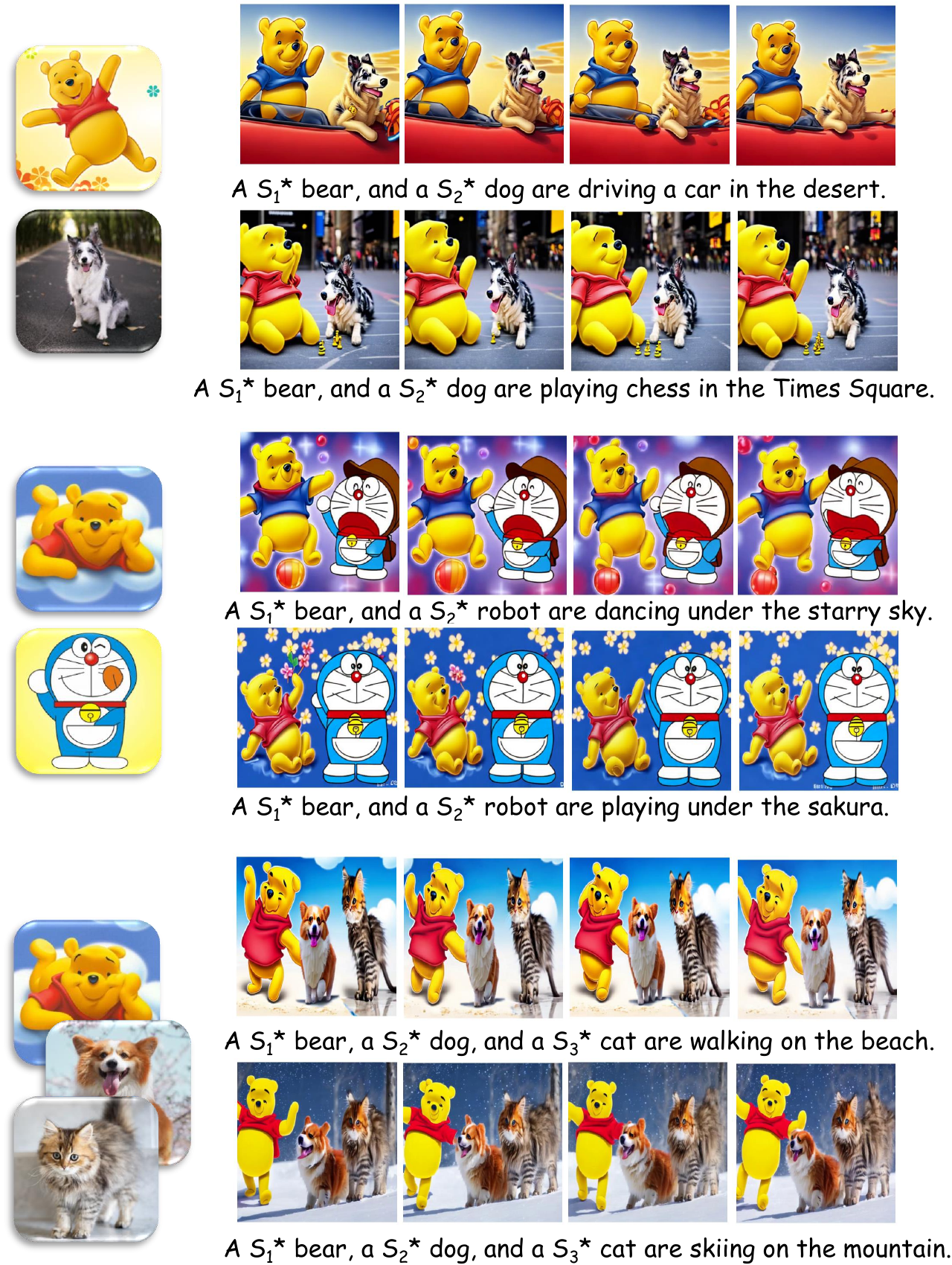}
    \caption{More generated cases with VideoDreamer.}
    \label{fig:more cases}
\end{figure}

\paragraph{More qualitative results} Besides the previously given qualitative examples, we provide more generated results on different customized subject customizations, where we put these subjects in different scenarios and make them conduct diverse actions. We provide the results in Figure~\ref{fig:more cases}.

\begin{figure}
    \centering
    \includegraphics[width=\linewidth]{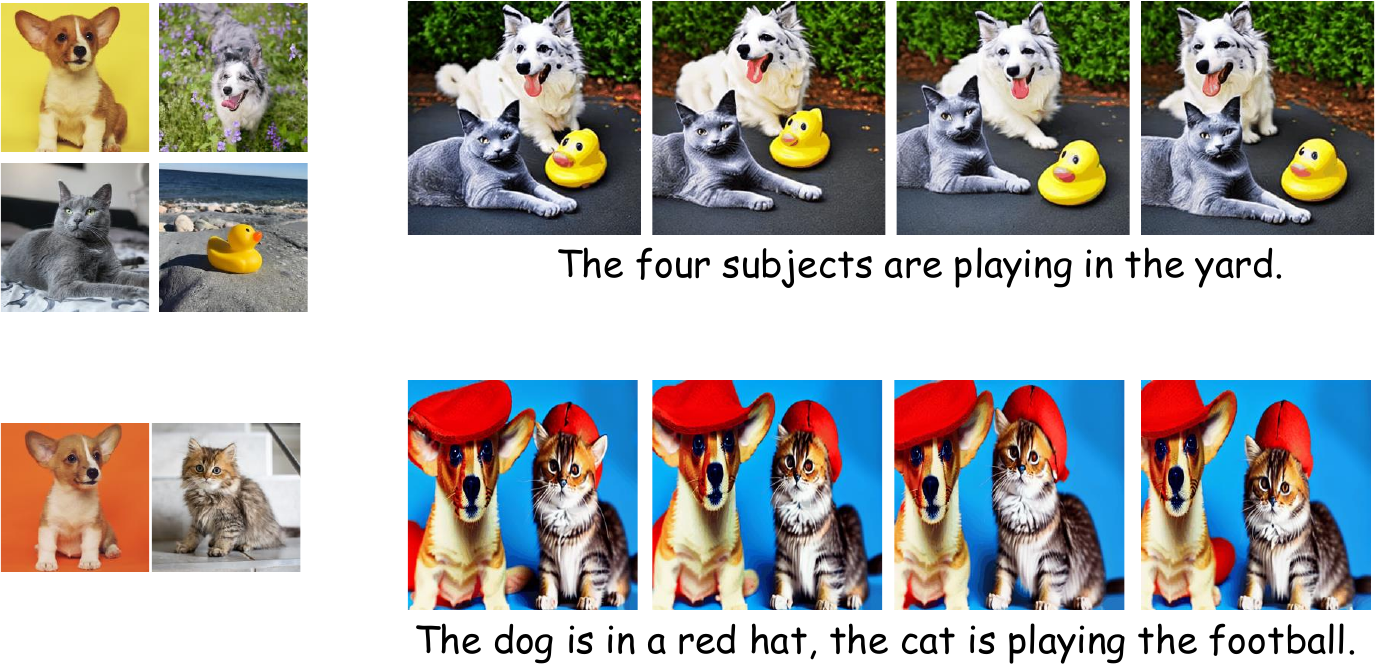}
    \caption{Failure generated cases.}
    \label{fig:failure}
\end{figure}

\paragraph{Failure cases} Although the proposed method is effective at generating videos for multiple customized subjects, we encountered some failure cases during the experiments. As shown in the first example in Fig.~\ref{fig:failure}, we try to apply our proposed method to 4-subject customization, but we find that in the generated videos, the first dog is missing. This phenomenon indicates that our although our proposed method works well for 2- or 3-subject combination, but its performance will drop when increasing the subject number. Additionally, our method faces the challenge of assigning specific attributes to each customized subject. As shown in the second example of Figure~\ref{fig:failure}, when we expect the dog to wear a red hat while the second cat to play football, both of them wear a red hat and no one plays the football. We hope future works can solve these problems.

\section{Limitation and Future Work}
Since this is the first attempt at customized multi-subject text-to-video generation, this work has some limitations. The first limitation is the evaluation benchmark. Although the MultiStudioBench evaluates the generation quality from comprehensive aspects, the data it applies is not large-scale and cannot cover all the varieties of subjects in the real world. In the future, we will enrich the benchmark with more diversified data. As for the method, the motion customization strategy currently can only be applied to the video generation model with decoupled spatial and temporal modules. Developing a general motion-customization finetuning approach could be an interesting future work. Additionally, we use~\cite{text2video0, guo2023animatediff} as the base video generator, where the single prompt is used to control all frames, making it hard to create videos with a dynamic background or multiple events, e.g., ``from the forest to the ocean'', and ``first play basketball and then dance''. How to tackle this problem is also worth exploring in the future.

\section{Conclusion}
\label{sec:conclusion}
In this paper, we present the first attempt at customized multi-subject text-to-video generation, and propose VideoDreamer, which can generate temporally consistent text-guided videos that faithfully preserve the subject identity, with the proposed Disen-Mix and HLR finetuning strategy. Extensive experiments on the proposed MultiStudioBench benchmark demonstrate that VideoDreamer has a remarkable ability in generating videos with new content for the given customized multiple subjects. Additionally, we provide an effective way to provide customized motion for the subjects. We believe this work takes a further step towards a more practical-to-used video generation system, and will inspire a lot of future works both in pretrained text-to-video models and finetuning methods.

\bibliographystyle{IEEEtran}
\bibliography{example_paper}

\begin{thebibliography}{10}
\providecommand{\url}[1]{#1}
\csname url@samestyle\endcsname
\providecommand{\newblock}{\relax}
\providecommand{\bibinfo}[2]{#2}
\providecommand{\BIBentrySTDinterwordspacing}{\spaceskip=0pt\relax}
\providecommand{\BIBentryALTinterwordstretchfactor}{4}
\providecommand{\BIBentryALTinterwordspacing}{\spaceskip=\fontdimen2\font plus
\BIBentryALTinterwordstretchfactor\fontdimen3\font minus \fontdimen4\font\relax}
\providecommand{\BIBforeignlanguage}[2]{{%
\expandafter\ifx\csname l@#1\endcsname\relax
\typeout{** WARNING: IEEEtran.bst: No hyphenation pattern has been}%
\typeout{** loaded for the language `#1'. Using the pattern for}%
\typeout{** the default language instead.}%
\else
\language=\csname l@#1\endcsname
\fi
#2}}
\providecommand{\BIBdecl}{\relax}
\BIBdecl

\bibitem{videodataset1}
M.~Bain, A.~Nagrani, G.~Varol, and A.~Zisserman, ``Frozen in time: A joint video and image encoder for end-to-end retrieval,'' in \emph{Proceedings of the IEEE/CVF International Conference on Computer Vision}, 2021, pp. 1728--1738.

\bibitem{videodataset2}
H.~Xue, T.~Hang, Y.~Zeng, Y.~Sun, B.~Liu, H.~Yang, J.~Fu, and B.~Guo, ``Advancing high-resolution video-language representation with large-scale video transcriptions,'' in \emph{Proceedings of the IEEE/CVF Conference on Computer Vision and Pattern Recognition}, 2022, pp. 5036--5045.

\bibitem{cogvideo}
W.~Hong, M.~Ding, W.~Zheng, X.~Liu, and J.~Tang, ``Cogvideo: Large-scale pretraining for text-to-video generation via transformers,'' in \emph{The Eleventh International Conference on Learning Representations}, 2022.

\bibitem{laion}
C.~Schuhmann, R.~Beaumont, R.~Vencu, C.~Gordon, R.~Wightman, M.~Cherti, T.~Coombes, A.~Katta, C.~Mullis, M.~Wortsman \emph{et~al.}, ``Laion-5b: An open large-scale dataset for training next generation image-text models,'' \emph{Advances in Neural Information Processing Systems}, vol.~35, pp. 25\,278--25\,294, 2022.

\bibitem{makeavideo}
U.~Singer, A.~Polyak, T.~Hayes, X.~Yin, J.~An, S.~Zhang, Q.~Hu, H.~Yang, O.~Ashual, O.~Gafni \emph{et~al.}, ``Make-a-video: Text-to-video generation without text-video data,'' in \emph{The Eleventh International Conference on Learning Representations}, 2023.

\bibitem{makeyourvideo}
J.~Xing, M.~Xia, Y.~Liu, Y.~Zhang, Y.~Zhang, Y.~He, H.~Liu, H.~Chen, X.~Cun, X.~Wang, Y.~Shan, and T.~Wong, ``Make-your-video: Customized video generation using textual and structural guidance,'' \emph{{IEEE} Trans. Vis. Comput. Graph.}, vol.~31, pp. 1526--1541, 2025.

\bibitem{text2video0}
L.~Khachatryan, A.~Movsisyan, V.~Tadevosyan, R.~Henschel, Z.~Wang, S.~Navasardyan, and H.~Shi, ``Text2video-zero: Text-to-image diffusion models are zero-shot video generators,'' in \emph{Proceedings of the IEEE/CVF International Conference on Computer Vision}, 2023, pp. 15\,954--15\,964.

\bibitem{free-bloom}
H.~Huang, Y.~Feng, C.~Shi, L.~Xu, J.~Yu, and S.~Yang, ``Free-bloom: Zero-shot text-to-video generator with llm director and ldm animator,'' \emph{Advances in Neural Information Processing Systems}, vol.~36, 2024.

\bibitem{imagenvideo}
J.~Ho, W.~Chan, C.~Saharia, J.~Whang, R.~Gao, A.~Gritsenko, D.~P. Kingma, B.~Poole, M.~Norouzi, D.~J. Fleet \emph{et~al.}, ``Imagen video: High definition video generation with diffusion models,'' \emph{arXiv preprint arXiv:2210.02303}, 2022.

\bibitem{magicvideo}
D.~Zhou, W.~Wang, H.~Yan, W.~Lv, Y.~Zhu, and J.~Feng, ``Magicvideo: Efficient video generation with latent diffusion models,'' \emph{arXiv preprint arXiv:2211.11018}, 2022.

\bibitem{guo2023animatediff}
Y.~Guo, C.~Yang, A.~Rao, Z.~Liang, Y.~Wang, Y.~Qiao, M.~Agrawala, D.~Lin, and B.~Dai, ``Animatediff: Animate your personalized text-to-image diffusion models without specific tuning,'' in \emph{The Twelfth International Conference on Learning Representations}, 2024.

\bibitem{Imagen}
C.~Saharia, W.~Chan, S.~Saxena, L.~Li, J.~Whang, E.~L. Denton, K.~Ghasemipour, R.~Gontijo~Lopes, B.~Karagol~Ayan, T.~Salimans \emph{et~al.}, ``Photorealistic text-to-image diffusion models with deep language understanding,'' \emph{Advances in Neural Information Processing Systems}, vol.~35, pp. 36\,479--36\,494, 2022.

\bibitem{dalle}
A.~Ramesh, M.~Pavlov, G.~Goh, S.~Gray, C.~Voss, A.~Radford, M.~Chen, and I.~Sutskever, ``Zero-shot text-to-image generation,'' in \emph{International Conference on Machine Learning}.\hskip 1em plus 0.5em minus 0.4em\relax PMLR, 2021, pp. 8821--8831.

\bibitem{glide}
A.~Q. Nichol, P.~Dhariwal, A.~Ramesh, P.~Shyam, P.~Mishkin, B.~Mcgrew, I.~Sutskever, and M.~Chen, ``Glide: Towards photorealistic image generation and editing with text-guided diffusion models,'' in \emph{International Conference on Machine Learning}.\hskip 1em plus 0.5em minus 0.4em\relax PMLR, 2022, pp. 16\,784--16\,804.

\bibitem{Sdiffusion}
R.~Rombach, A.~Blattmann, D.~Lorenz, P.~Esser, and B.~Ommer, ``High-resolution image synthesis with latent diffusion models,'' in \emph{Proceedings of the IEEE/CVF Conference on Computer Vision and Pattern Recognition}, 2022, pp. 10\,684--10\,695.

\bibitem{dalle2}
A.~Ramesh, P.~Dhariwal, A.~Nichol, C.~Chu, and M.~Chen, ``Hierarchical text-conditional image generation with clip latents,'' \emph{arXiv preprint arXiv:2204.06125}, 2022.

\bibitem{earlyt2v1}
Y.~Li, M.~Min, D.~Shen, D.~Carlson, and L.~Carin, ``Video generation from text,'' in \emph{Proceedings of the AAAI conference on artificial intelligence}, vol.~32, no.~1, 2018.

\bibitem{earlyt2v2}
Y.~Liu, X.~Wang, Y.~Yuan, and W.~Zhu, ``Cross-modal dual learning for sentence-to-video generation,'' in \emph{Proceedings of the 27th ACM international conference on multimedia}, 2019, pp. 1239--1247.

\bibitem{earlyt2v3}
T.~Marwah, G.~Mittal, and V.~N. Balasubramanian, ``Attentive semantic video generation using captions,'' in \emph{Proceedings of the IEEE international conference on computer vision}, 2017, pp. 1426--1434.

\bibitem{earlyt2v4}
G.~Mittal, T.~Marwah, and V.~N. Balasubramanian, ``Sync-draw: Automatic video generation using deep recurrent attentive architectures,'' in \emph{Proceedings of the 25th ACM international conference on Multimedia}, 2017, pp. 1096--1104.

\bibitem{latentvideo}
Y.~He, T.~Yang, Y.~Zhang, Y.~Shan, and Q.~Chen, ``Latent video diffusion models for high-fidelity video generation with arbitrary lengths,'' \emph{arXiv preprint arXiv:2211.13221}, 2022.

\bibitem{videofusion}
Z.~Luo, D.~Chen, Y.~Zhang, Y.~Huang, L.~Wang, Y.~Shen, D.~Zhao, J.~Zhou, and T.~Tan, ``Videofusion: Decomposed diffusion models for high-quality video generation,'' in \emph{Proceedings of the IEEE/CVF Conference on Computer Vision and Pattern Recognition}, 2023, pp. 10\,209--10\,218.

\bibitem{phenaki}
R.~Villegas, M.~Babaeizadeh, P.-J. Kindermans, H.~Moraldo, H.~Zhang, M.~T. Saffar, S.~Castro, J.~Kunze, and D.~Erhan, ``Phenaki: Variable length video generation from open domain textual descriptions,'' in \emph{The Eleventh International Conference on Learning Representations}, 2023.

\bibitem{godiva}
C.~Wu, L.~Huang, Q.~Zhang, B.~Li, L.~Ji, F.~Yang, G.~Sapiro, and N.~Duan, ``Godiva: Generating open-domain videos from natural descriptions,'' \emph{arXiv preprint arXiv:2104.14806}, 2021.

\bibitem{video-p2p}
S.~Liu, Y.~Zhang, W.~Li, Z.~Lin, and J.~Jia, ``Video-p2p: Video editing with cross-attention control,'' in \emph{2024 IEEE/CVF Conference on Computer Vision and Pattern Recognition (CVPR)}.\hskip 1em plus 0.5em minus 0.4em\relax IEEE, 2024, pp. 8599--8608.

\bibitem{tune-a-video}
J.~Z. Wu, Y.~Ge, X.~Wang, S.~W. Lei, Y.~Gu, Y.~Shi, W.~Hsu, Y.~Shan, X.~Qie, and M.~Z. Shou, ``Tune-a-video: One-shot tuning of image diffusion models for text-to-video generation,'' in \emph{Proceedings of the IEEE/CVF International Conference on Computer Vision}, 2023, pp. 7623--7633.

\bibitem{vid2vid-0}
W.~Wang, K.~Xie, Z.~Liu, H.~Chen, Y.~Cao, X.~Wang, and C.~Shen, ``Zero-shot video editing using off-the-shelf image diffusion models,'' \emph{arXiv preprint arXiv:2303.17599}, 2023.

\bibitem{controlvideo}
M.~Zhao, R.~Wang, F.~Bao, C.~Li, and J.~Zhu, ``Controlvideo: Adding conditional control for one shot text-to-video editing,'' \emph{arXiv preprint arXiv:2305.17098}, 2023.

\bibitem{dreamix}
E.~Molad, E.~Horwitz, D.~Valevski, A.~R. Acha, Y.~Matias, Y.~Pritch, Y.~Leviathan, and Y.~Hoshen, ``Dreamix: Video diffusion models are general video editors,'' \emph{arXiv preprint arXiv:2302.01329}, 2023.

\bibitem{render-a-video}
S.~Yang, Y.~Zhou, Z.~Liu, and C.~C. Loy, ``Rerender a video: Zero-shot text-guided video-to-video translation,'' in \emph{SIGGRAPH Asia 2023 Conference Papers}, 2023, pp. 1--11.

\bibitem{fatezero}
C.~Qi, X.~Cun, Y.~Zhang, C.~Lei, X.~Wang, Y.~Shan, and Q.~Chen, ``Fatezero: Fusing attentions for zero-shot text-based video editing,'' in \emph{Proceedings of the IEEE/CVF International Conference on Computer Vision}, 2023, pp. 15\,932--15\,942.

\bibitem{dreambooth}
N.~Ruiz, Y.~Li, V.~Jampani, Y.~Pritch, M.~Rubinstein, and K.~Aberman, ``Dreambooth: Fine tuning text-to-image diffusion models for subject-driven generation,'' in \emph{Proceedings of the IEEE/CVF Conference on Computer Vision and Pattern Recognition}, 2023, pp. 22\,500--22\,510.

\bibitem{TI}
R.~Gal, Y.~Alaluf, Y.~Atzmon, O.~Patashnik, A.~H. Bermano, G.~Chechik, and D.~Cohen-or, ``An image is worth one word: Personalizing text-to-image generation using textual inversion,'' in \emph{The Eleventh International Conference on Learning Representations}, 2023.

\bibitem{customfusion}
N.~Kumari, B.~Zhang, R.~Zhang, E.~Shechtman, and J.-Y. Zhu, ``Multi-concept customization of text-to-image diffusion,'' in \emph{Proceedings of the IEEE/CVF Conference on Computer Vision and Pattern Recognition}, 2023, pp. 1931--1941.

\bibitem{svdiff}
L.~Han, Y.~Li, H.~Zhang, P.~Milanfar, D.~Metaxas, and F.~Yang, ``Svdiff: Compact parameter space for diffusion fine-tuning,'' in \emph{Proceedings of the IEEE/CVF International Conference on Computer Vision}, 2023, pp. 7323--7334.

\bibitem{disenbooth}
H.~Chen, Y.~Zhang, S.~Wu, X.~Wang, X.~Duan, Y.~Zhou, and W.~Zhu, ``Disenbooth: Identity-preserving disentangled tuning for subject-driven text-to-image generation,'' in \emph{The Twelfth International Conference on Learning Representations}, 2024.

\bibitem{mix-of-show}
Y.~Gu, X.~Wang, J.~Z. Wu, Y.~Shi, Y.~Chen, Z.~Fan, W.~Xiao, R.~Zhao, S.~Chang, W.~Wu \emph{et~al.}, ``Mix-of-show: Decentralized low-rank adaptation for multi-concept customization of diffusion models,'' \emph{Advances in Neural Information Processing Systems}, vol.~36, pp. 15\,890--15\,902, 2023.

\bibitem{elite}
Y.~Wei, Y.~Zhang, Z.~Ji, J.~Bai, L.~Zhang, and W.~Zuo, ``Elite: Encoding visual concepts into textual embeddings for customized text-to-image generation,'' in \emph{Proceedings of the IEEE/CVF International Conference on Computer Vision}, 2023, pp. 15\,943--15\,953.

\bibitem{Aslearning}
W.~Chen, H.~Hu, Y.~Li, N.~Ruiz, X.~Jia, M.-W. Chang, and W.~W. Cohen, ``Subject-driven text-to-image generation via apprenticeship learning,'' \emph{Advances in Neural Information Processing Systems}, vol.~36, pp. 30\,286--30\,305, 2023.

\bibitem{Instantbooth}
J.~Shi, W.~Xiong, Z.~Lin, and H.~J. Jung, ``Instantbooth: Personalized text-to-image generation without test-time finetuning,'' in \emph{Proceedings of the IEEE/CVF conference on computer vision and pattern recognition}, 2024, pp. 8543--8552.

\bibitem{fastcomposer}
G.~Xiao, T.~Yin, W.~T. Freeman, F.~Durand, and S.~Han, ``Fastcomposer: Tuning-free multi-subject image generation with localized attention,'' \emph{International Journal of Computer Vision}, pp. 1--20, 2024.

\bibitem{subject-diffusion}
J.~Ma, J.~Liang, C.~Chen, and H.~Lu, ``Subject-diffusion: Open domain personalized text-to-image generation without test-time fine-tuning,'' in \emph{ACM SIGGRAPH 2024 Conference Papers}, 2024, pp. 1--12.

\bibitem{break-a-scene}
O.~Avrahami, K.~Aberman, O.~Fried, D.~Cohen-Or, and D.~Lischinski, ``Break-a-scene: Extracting multiple concepts from a single image,'' in \emph{SIGGRAPH Asia 2023 Conference Papers}, 2023, pp. 1--12.

\bibitem{videoassembler}
H.~Zhao, T.~Lu, J.~Gu, X.~Zhang, Z.~Wu, H.~Xu, and Y.-G. Jiang, ``Videoassembler: Identity-consistent video generation with reference entities using diffusion model,'' \emph{arXiv preprint arXiv:2311.17338}, 2023.

\bibitem{videobooth}
Y.~Jiang, T.~Wu, S.~Yang, C.~Si, D.~Lin, Y.~Qiao, C.~C. Loy, and Z.~Liu, ``Videobooth: Diffusion-based video generation with image prompts,'' in \emph{Proceedings of the IEEE/CVF Conference on Computer Vision and Pattern Recognition}, 2024, pp. 6689--6700.

\bibitem{dreamvideo}
Y.~Wei, S.~Zhang, Z.~Qing, H.~Yuan, Z.~Liu, Y.~Liu, Y.~Zhang, J.~Zhou, and H.~Shan, ``Dreamvideo: Composing your dream videos with customized subject and motion,'' in \emph{Proceedings of the IEEE/CVF Conference on Computer Vision and Pattern Recognition}, 2024, pp. 6537--6549.

\bibitem{wang2024visual}
W.~Wang, Y.~Yang, and Y.~Pan, ``Visual knowledge in the big model era: Retrospect and prospect,'' \emph{Frontiers of Information Technology \& Electronic Engineering}, vol.~26, no.~1, pp. 1--19, 2025.

\bibitem{Unet}
O.~Ronneberger, P.~Fischer, and T.~Brox, ``U-net: Convolutional networks for biomedical image segmentation,'' in \emph{Medical Image Computing and Computer-Assisted Intervention--MICCAI 2015: 18th International Conference, Munich, Germany, October 5-9, 2015, Proceedings, Part III 18}.\hskip 1em plus 0.5em minus 0.4em\relax Springer, 2015, pp. 234--241.

\bibitem{DDPM}
J.~Ho, A.~Jain, and P.~Abbeel, ``Denoising diffusion probabilistic models,'' \emph{Advances in neural information processing systems}, vol.~33, pp. 6840--6851, 2020.

\bibitem{DDIM}
J.~Song, C.~Meng, and S.~Ermon, ``Denoising diffusion implicit models,'' in \emph{International Conference on Learning Representations}, 2020.

\bibitem{DPM}
C.~Lu, Y.~Zhou, F.~Bao, J.~Chen, C.~Li, and J.~Zhu, ``Dpm-solver: A fast ode solver for diffusion probabilistic model sampling in around 10 steps,'' \emph{Advances in Neural Information Processing Systems}, vol.~35, pp. 5775--5787, 2022.

\bibitem{attr}
W.~Feng, X.~He, T.-J. Fu, V.~Jampani, A.~R. Akula, P.~Narayana, S.~Basu, X.~E. Wang, and W.~Y. Wang, ``Training-free structured diffusion guidance for compositional text-to-image synthesis,'' in \emph{The Eleventh International Conference on Learning Representations}, 2022.

\bibitem{LoRA}
E.~J. Hu, P.~Wallis, Z.~Allen-Zhu, Y.~Li, S.~Wang, L.~Wang, W.~Chen \emph{et~al.}, ``Lora: Low-rank adaptation of large language models,'' in \emph{International Conference on Learning Representations}.

\bibitem{hugging}
P.~V. Platen, S.~Patil, A.~Lozhkov, P.~Cuenca, N.~Lambert, K.~Rasul, M.~Davaadorj, and T.~Wolf, ``Diffusers: State-of-the-art diffusion models,'' \url{https://github.com/huggingface/diffusers}, 2022.

\bibitem{adamw}
I.~Loshchilov and F.~Hutter, ``Decoupled weight decay regularization,'' in \emph{International Conference on Learning Representations}.

\bibitem{pytorch}
A.~Paszke, S.~Gross, F.~Massa, A.~Lerer, J.~Bradbury, G.~Chanan, T.~Killeen, Z.~Lin, N.~Gimelshein, L.~Antiga \emph{et~al.}, ``Pytorch: An imperative style, high-performance deep learning library,'' \emph{Advances in neural information processing systems}, vol.~32, 2019.

\bibitem{opencv}
G.~Bradski, ``{The OpenCV Library},'' \emph{Dr. Dobb's Journal of Software Tools}, 2000.

\bibitem{vbench}
Z.~Huang, Y.~He, J.~Yu, F.~Zhang, C.~Si, Y.~Jiang, Y.~Zhang, T.~Wu, Q.~Jin, N.~Chanpaisit \emph{et~al.}, ``Vbench: Comprehensive benchmark suite for video generative models,'' in \emph{Proceedings of the IEEE/CVF Conference on Computer Vision and Pattern Recognition}, 2024, pp. 21\,807--21\,818.

\end{thebibliography}

\end{document}